\documentclass[acmsmall]{acmart}
\setcopyright{acmcopyright}
\copyrightyear{2023}
\acmYear{2023}
\acmDOI{10.1145/XXXXXX}

\acmJournal{TIST}
\acmVolume{1}   
\acmNumber{1}
\acmArticle{1}
\acmMonth{7}

\usepackage{amsmath,amsfonts}
\usepackage{euler}
\usepackage{algorithmic}
\usepackage{graphicx}
\usepackage{textcomp}
\usepackage{xcolor}
\usepackage{xspace}
\usepackage[ruled, vlined, linesnumbered]{algorithm2e}
\usepackage{algorithmic}
\usepackage{graphicx}
\usepackage{subcaption}
\usepackage{amsthm}
\usepackage{mathrsfs}
\usepackage{booktabs}
\usepackage{color}

\newcommand{\ourm}{\textsc{ProActive}\xspace}
\newcommand{\ourp}{\textsc{ProActive++}\xspace}
\newcommand{\ours}{\textsc{ProAct}\xspace}

\newcommand{\cm}[1]{\mathcal{#1}}

\newcommand{\bs}[1]{\boldsymbol{#1}}
\newcommand{\eg}{\emph{e.g.}}
\newcommand{\ie}{\emph{i.e.}}
\newcommand{\etc}{\emph{etc.}}
\newcommand{\cat}[1]{`\textit{#1}'}

\newcommand{\xhdr}[1]{\vspace{1mm} \noindent{{\bf #1.}}}

\newcommand{\bfast}{Breakfast}
\newcommand{\mult}{Multi-THUMOS}
\newcommand{\act}{Activity-Net}
\newcommand{\actnew}{Activity-Sparse}

\theoremstyle{definition}
\newtheorem{definition}{Definition}
\newtheorem*{problem*}{Problem Statement}

\renewcommand{\comment}[1]{}
\usepackage{xcolor}
\usepackage{multirow,amsmath,booktabs,dcolumn}
\usepackage{todonotes}
\usepackage{colortbl}
\usepackage{paralist}
\usepackage{graphicx}
\usepackage{caption}
\usepackage{lipsum}
\usepackage{mathtools}

\begin{document}
\title{Tapestry of Time and Actions: Modeling Human Activity Sequences using Temporal Point Process Flows}
\titlenote{This is a substantially refined and expanded version of \citet{proactive} that appeared in the Proc. of KDD 2022.}

\author{Vinayak Gupta}
\affiliation{
  \institution{University of Washington Seattle}
  \city{Seattle, Washington}
  \country{United States}}
  \email{vinayak@cs.washington.edu}

\author{Srikanta Bedathur}
\affiliation{
  \institution{Indian Institute of Technology Delhi}
  \city{New Delhi}
  \country{India}}
  \email{srikanta@cse.iitd.ac.in}

\renewcommand{\shortauthors}{Gupta et al.}
\renewcommand{\shorttitle}{Point Processes for Action Sequences}

\begin{abstract}
Human beings always engage in a vast range of activities and tasks that demonstrate their ability to adapt to different scenarios. These activities can range from the simplest daily routines, like walking and sitting, to multi-level complex endeavors such as cooking a four-course meal.  Any human activity can be represented as a temporal sequence of actions performed to achieve a certain goal. Unlike the time series datasets extracted from electronics or machines, these action sequences are highly disparate in their nature -- the time to finish a sequence of actions can vary between different persons. Therefore, understanding the dynamics of these sequences is essential for many downstream tasks such as activity length prediction, goal prediction, next-action recommendation, \etc\ Existing neural network-based approaches that learn a continuous-time activity sequence (or CTAS) are limited to the presence of only visual data or are designed specifically for a particular task, \ie, limited to next action or goal prediction. In this paper, we present \ourm, a neural marked temporal point process (MTPP) framework for modeling the continuous-time distribution of actions in an activity sequence while simultaneously addressing three high-impact problems -- next action prediction, sequence-goal prediction, and \textit{end-to-end} sequence generation. Specifically, we utilize a self-attention module with temporal normalizing flows to model the influence and the inter-arrival times between actions in a sequence. Moreover, for time-sensitive prediction, we perform an \textit{early} detection of sequence goal via a constrained margin-based optimization procedure. This in-turn allows \ourm to predict the sequence goal using a limited number of actions. In addition, we propose a novel addition over the \ourm model that can handle variations in the order of actions, \ie, different methods of achieving a given goal. We demonstrate that this variant can learn the order in which the person or actor prefers to do their actions. Extensive experiments on sequences derived from three activity recognition datasets show the significant accuracy boost of \ourm over the state-of-the-art in terms of action and goal prediction, and the first-ever application of end-to-end action sequence generation.
\end{abstract}


\maketitle

\section{Introduction}\label{sec:intro}
Human beings possess a remarkable capacity for adaptation, enabling them to engage in an extensive array of activities and tasks across various domains. These activities encompass a wide spectrum, ranging from the mundane routines of daily life to intricate endeavors that require a high level of skill and coordination. At the simplest level, humans perform basic actions like walking and sitting as part of their everyday routines. On a more complex level, humans engage in multifaceted tasks that require planning, problem-solving, and the integration of various skills. Consider the process of cooking a four-course meal. It involves a sequence of actions such as prepping ingredients, following recipes, coordinating cooking times, and plating the dishes in a harmonious manner. Each step builds upon the previous one, leading to the ultimate goal of preparing a delicious and satisfying meal. Recognizing that any human activity can be represented as a temporal sequence of actions opens up avenues for understanding and analyzing behavior. A majority of the data generated via human activities, \eg, running, playing basketball, cooking, \etc, can be represented as a sequence of actions over a continuous time. These actions denote a step taken by an actor\footnote{We use \textit{actor} to refer to the person performing actions in a sequence.} towards achieving a certain goal and vary in their start and completion times, depending on the actor and the surrounding environment\cite{mehrasa2017learning,breakfast,avae}. Therefore, unlike synthetic time series, these continuous-time action sequences (or CTAS) can vary significantly even if they consist of the same set of actions. For \eg, one person making omelets may take a longer time to cook eggs while another may prefer to cook for a short time\footnote{\url{https://bit.ly/3F5aEwX} (Accessed May 2023)}; or in a football game, Xavi may make a faster pass than Pirlo, even though the goals and the sequence of actions are the same. In addition, modeling the dynamics of CTAS becomes increasingly challenging due to the limited ability of the current neural frameworks, recurrent or self-attention-based, in capturing the continuous nature of action times~\cite{transformer,sasrec}. This situation has been further exacerbated due to the large variance in action-\textit{times} and \textit{types} of the actions a person can perform. Therefore, the problem of modeling a CTAS has been overlooked by the past literature.

In recent years, neural marked temporal point processes (MTPP) have shown significant promise in modeling a variety of continuous-time sequences in healthcare~\cite{rizoiu2,rizoiu1,neuroseqret}, finance~\cite{sahp,bacry}, education~\cite{sahebi}, and social networks~\cite{leskovec,nhp,thp}. However, standard MTPP have a limited modeling ability for CTAS as: (i) they assume homogeneity among sequences, \ie, they cannot distinguish between two sequences of similar actions but with different time duration; (ii) in a CTAS, an action may finish before the start of the next action and thus, to model this empty time interval an MTPP must introduce a new action type, \eg, \textit{NULL} or \textit{end-action}, which may lead to an unwarranted increase in the types of actions to be modeled; and (iii) they cannot encapsulate the additional features associated with an action, for \eg, minimum time for completion, necessary previous actions, or can be extended to sequence generation.

\vspace{-0.05cm}
\subsection{Our Contribution}
In this paper, we present \ourm (\textbf{P}oint P\textbf{ro}cess flows for \textbf{Activ}ity S\textbf{e}quences), a normalizing flow-based neural MTPP framework, designed specifically to model the dynamics of a CTAS. Specifically, \ourm addresses three challenging problems -- (i) action prediction; (ii) goal detection; and (iii) the first-of-its-kind task of \textit{end-to-end} sequence generation. We learn the distribution of actions in a CTAS using temporal normalizing flows (NF)~\cite{shakir,ppflows} conditioned on the dynamics of the sequence as well as the action features (\eg, minimum completion time, \etc). Such a flow-based formulation provides \ourm flexibility over other similar frameworks~\cite{karishma,intfree} to better model the inter-action dynamics within and across the sequences. Moreover, our model is designed for \textit{early} detection of the sequence goal, \ie, identifying the result of a CTAS without traversing the complete sequence. We achieve this by using a time-bounded optimization procedure, \ie, by incrementally increasing the probability of identifying the true goal via a \textit{margin}-based and a weighted factor-based learning~\cite{margin,earlyijcv}. Such an optimization procedure allows \ourm to model the goal-action hierarchy within a sequence, \ie, the \textit{necessary} set of actions towards achieving a particular goal, and simultaneously, the order of actions within CTAS with similar goals.

 \ourm is designed to effectively perform three real-world tasks -- action prediction, goal detection, and end-to-end CTAS generation. Specifically, the task of action prediction involves predicting the actions that a person might perform in the future towards achieving a specific goal. Goal detection involves identifying and recognizing the goals or the \textit{end-intention} of a person given their past actions. Lastly, our model demonstrates remarkable performance for end-to-end CTAS generation. CTAS generation refers to the task of generating coherent and meaningful sequences of actions on a given sequence goal as input. This indicates that our model has a better grasp of the underlying structure and dependencies within the CTAS data, allowing it to generate more plausible and accurate outcomes. To the best of our knowledge, we present the first-ever application of MTPP via \textit{end-to-end} action sequence generation. Such a novel ability for MTPP models can reinforce their usage in applications related to bio-signals~\cite{haradal}, sensor-data~\cite{sensegen}, \etc, and overcome the modeling challenge due to scarcity of activity data~\cite{donahue2020user,luo2020database}. Buoyed by the success of attention models in sequential applications~\cite{transformer}, we use a self-attention architecture in \ourm to model the inter-action influences in a CTAS. In summary, the key contributions we make in this paper are:
\begin{itemize}
\item We propose \ourm, a novel temporal flow-based MTPP framework designed specifically for modeling human activities with a time-bounded optimization framework for early detection of CTAS goal. 
\item Our normalizing flow-based modeling framework incorporates the sequence and individual action dynamics along with the action-goal hierarchy. Thus, \ourm introduces the first-of-its-kind MTPP application of end-to-end action CTAS generation with just the sequence goal as input.
\item Empirical results show that \ourm outperforms the state-of-the-art models for all three tasks --  action prediction, goal detection, and sequence generation.
\item We present a novel variant of \ourm that can handle differences between CTAS that have a similar goal, which makes it more suitable for predicting future actions.  
\end{itemize}

\subsection{Organization}
We present a problem formulation and a background on necessary techniques in Section~\ref{sec:basics} and Section~\ref{sec:psetup} respectively. Section~\ref{sec:model} gives an overview followed by a detailed development of all components in \ourm. Section~\ref{sec:exp} contains the in-depth experimental analysis, and Section~\ref{sec:related} has a comprehensive review of relevant works in the field, examining prior research and approaches that are closely related to the topic of this paper, before concluding our discussion in Section~\ref{sec:conc}.

\section{Background} \label{sec:basics}
In this section, we provide the essential background information for this paper. Specifically, we introduce the concepts of Marked Temporal Point Processes (MTPP) and Normalizing Flows (NF). These comncepts form the foundational knowledge required to grasp the core ideas of \ourm.

\subsection{Marked Temporal Point Processes}
MTPP\cite{hawkes} are probabilistic generative models for continuous-time event sequences. An MTPP can be represented as a probability distribution over sequences of variable length belonging to a time interval $[0, T]$. Equivalently, they can be described using a counting process, say $N(t)$, and are characterized by the underlying conditional intensity function, $\lambda^*(t)$ which specifies the likelihood of the next event, conditioned on the history of events. The intensity function $\lambda^*(t)$ computes the infinitesimal probability that an event will happen in the time window $(t, t + dt]$ conditioned on the history as:
\begin{equation}
\mathbb{P} \big(d N(t) = N(t+dt) - N(t) = 1 \big) = \lambda^*(t),
\end{equation}
Here, $*$ denotes a dependence on history. Given the conditional intensity function, we obtain the probability density function as:
\begin{equation}
p^*(\Delta_{t, i}) = \lambda^*(t_{i-1} + \Delta_{t, i}) \exp \bigg(-\int_{0}^{\Delta_{t, i}} \lambda^*(t_{i-1} + r) dr\bigg),
\end{equation}
where, $\Delta_{t, i}$ denotes the inter-event time interval, \ie, $t_i - t_{i-1}$. In contrast to other neural MTPP models that rely on the intensity function~\cite{rmtpp,nhp,sahp,thp} we replace the intensity function with a \emph{lognormal} flow. Such a formulation facilitates closed-form and faster sampling as well as a more accurate prediction than the intensity-based models~\cite{intfree,ppflows}.

\subsection{Normalizing Flows}
Normalizing flows~\cite{shakir,ppflows} (NF) are generative models used for density estimation and event sampling. They work by mapping simple distributions to complex ones using multiple bijective, \ie, reversible functions. For \eg, the function $r(x)=x+1$ is a reversible function because, for each input, a unique output exists and vice-versa, whereas the function $r(x) = x^2$ is not a reversible function. In detail, let $\bs{Z} \in \mathbb{R}^D$ be a random variable with a known probability density function $p_{\bs{Z}} : \mathbb{R}^D \rightarrow \mathbb{R}$. Let $g$ be an invertible function and $\bs{Y} = g(\bs{Z})$. Then, via the change of variables formula~\cite{kobyzev2020normalizing}, the probability density function of $\bs{Y}$ is:
\begin{equation}
p_{\bs{Y}} (\bs{y}) = p_{\bs{Z}} \big( f(\bs{y}) \big) \, \Big | \mathrm{det} \, D f(\bs{y}) \Big |,
\end{equation}
where $f(\cdot)$ is the inverse of $g$ and $D f(\bs{y}) = \frac{\delta f}{\delta \bs{y}}$ is the Jacobian of $f$. Here, the above function $g(\cdot)$ (a generator) projects the base density $p(\bs{Z})$ to a more complex density, and this projection is considered to be in the \textit{generative} direction. Whereas the inverse function $f(\cdot)$ moves from a complicated distribution towards the simpler one of $p(\bs{Z})$, referred to as the \textit{normalizing} direction. Since in generative models, the base density $p(\bs{Z})$ is considered as Normal distribution, this formulation gives rise to the name \textit{normalizing} flows. To sample a point $\bs{y}$, one can sample a point $\bs{z}$ and then apply the generator $\bs{y} = g(\bs{z})$. Such a procedure supports closed-form sampling. Moreover, modern approaches for normalizing flows approximate the above functions using a neural network~\cite{autoregressive,dhaliwal,shakir}. Normalizing flows have been increasingly used to define flexible and theoretically sound models for marked temporal point processes~\cite{intfree, ppflows}.

\section{Problem Formulation} \label{sec:psetup}
As mentioned in Section~\ref{sec:intro}, we represent an activity via a continuous-time action sequence, \ie, a series of actions undertaken by actors and their corresponding time of occurrences. We derive each CTAS from annotated frames of videos consisting of individuals performing certain activities. Specifically, for every video, we have a sequence of activity labels being performed in the video along with timestamps for each activity. Therefore, each CTAS used in our dataset is derived from these sequences of a video. Formally, we provide a detailed description of a CTAS in Definition~\ref{def:ctas}.

\begin{definition}[Continuous Time Action Sequence]
\label{def:ctas}
\textit{We define a continuous-time action sequence (CTAS) as a series of action events taken by an actor to achieve a particular goal. Specifically, we represent a CTAS as $\cm{S}_k=\{e_i=(c_i, t_i) | i \in[k] , t_i<t_{i+1}\}$, where $t_i \in \mathbb{R}^+$ is the start-time of the action, $c_i\in \cm{C}$ is the discrete category or mark of the $i$-th action, $\cm{C}$ is the set of all categories, $\Delta_{t, i} = t_i - t_{i-1}$ as the inter-action time, and $\cm{S}_k$ denotes the sequence of first $k$ actions. Each CTAS has an associated result, $g \in \cm{G}$, that signifies the goal of the CTAS. Here, $\cm{G}$ denotes the set of all possible sequence goals.}
\end{definition}

To highlight the relationship between sequence goal and actions consider the example of a CTAS with the goal of \cat{making-coffee}, would comprise of actions -- \cat{take-a-cup}, \cat{pour-milk}, \cat{add-coffee-powder}, \cat{add-sugar}, and \cat{stir} -- at different time intervals. Given the aforementioned definitions, we formulate the tasks of the next action prediction, sequence goal detection, and sequence generation as:

\xhdr{Input}
A sequence of time-stamped actions done by a person, denoted by $\cm{S}_k$, consisting of -- type of action and starting times --  that lead to a specific goal, denoted by $g$.

\xhdr{Output}
A probabilistic prediction model that can present results for three distinct tasks -- (i) estimate the likelihood of the next action $e_{k+1}$ along with the action category and occurrence time; (ii) to predict the end-goal of the CTAS being modeled, \ie, $\widehat{g}$; and (iii) a generative model to sample a sequence of actions, $\widehat{\cm{S}}$ given the true sequence goal, $g$.

\section{Model}\label{sec:model}
In this section, we first present a high-level overview of the \ourm model and then describe the neural parameterization of each component in detail. Lastly, we provide a detailed description of its optimization and sequence generation procedure.

\subsection{High Level Overview}\label{sec:overview}
We use an MTPP denoted by $p_{\theta}(\cdot)$, to learn the generative mechanism of a continuous-time action sequence. Moreover, we design the sequence modeling framework of $p_{\theta}(\cdot)$ using a self-attention-based encoder-decoder model~\cite{transformer}. Specifically, we embed the actions in a CTAS, \ie, $\cm{S}_k$, to a vector embedding, denoted by $\bs{s}_k$, using a weighted aggregation of all past actions. Therefore, $\bs{s}_k$ signifies a compact neural representation of the sequence history, \ie, all actions till the $k$-th index and their marks and occurrence times. Recent research~\cite{thp,sahp} has shown that an attention-based modeling choice can better capture the long-term dependencies as compared to RNN-based MTPP models~\cite{rmtpp,nhp,intfree,fullyneural}. A detailed description of the embedding procedure is given in Section~\ref{sec:detail}. We use our MTPP $p_{\theta}(\cdot)$ to estimate the generative model for the $(k+1)$-th action conditioned on the past, \ie, $p(e_{k+1})$ as:
\begin{equation}
p_{\theta}(e_{k+1} | \bs{s}_k) = \mathbb{P}_{\theta}(c_{k+1}|\bs{s}_k) \cdot \rho_{\theta}(\Delta_{t, k+1}|\bs{s}_k),
\end{equation}
where $\mathbb{P}_{\theta}(\cdot)$ and $\rho_{\theta}(\cdot)$ denote the probability distribution of marks and the density function for inter-action arrival times respectively. Note that both the functions are conditioned on $\bs{s}_k$ and thus \ourm requires a joint optimizing procedure for both -- action time and mark prediction. Next, we describe the mechanism used in \ourm to predict the next action and goal detection in a CTAS.

\xhdr{Next Action Prediction}
We determine the most probable mark and time of the next action, using $p_{\theta} (\cdot)$ via standard sampling techniques over $\mathbb{P}_{\theta}(\cdot)$ and $\rho_{\theta}(\cdot)$ respectively~\cite{rmtpp,intfree}.
\begin{equation}
\widehat{e_{k+1}} \sim p_{\theta}(e_{k+1} | \bs{s}_k),
\end{equation}
In addition, to keep the history embedding up-to-date with the all past actions, we incrementally update $\bs{s}_k$ to $\bs{s}_{k+1}$ by incorporating the details of action $e_{k+1}$.

\xhdr{Goal Detection}
Since the history embedding, $\bs{s}_k$, represents an aggregation of all past actions in a sequence, it can also be used to capture the influences between actions and thus, can be extended to detect the goal of the CTAS. Specifically, to detect the CTAS goal, we use a non-linear transformation over $\bs{s}_k$ as:
\begin{equation}
\widehat{g} \sim \mathbb{P}_{g' \in \cm{G}}(\Phi(s_k)),
\label{eq:goal}
\end{equation}
where, $\mathbb{P}_{\bullet}$ denotes the distribution over all sequence goals and $\Phi(\cdot)$ denotes the transformation via a fully-connected MLP layer.

\subsection{Neural Parameterization}\label{sec:detail}
Here, we present a detailed description of the neural architecture of our MTPP, $p_{\theta}(\cdot)$, and the optimization procedure in \ourm. Specifically, we realize $p_{\theta}(\cdot)$ using a three-layer architecture:

\xhdr{Input Layer}
As mentioned in Section~\ref{sec:basics}, each action $e_i \in \cm{S}_k$ is represented by a mark $c_i$ and time $t_i$. Therefore, we embed each action as a combination of all these features as follows:
\begin{equation}
\bs{y}_i = \bs{w}_{y, c} c_i + \bs{w}_{y, t} t_{i} + \bs{w}_{y, \Delta} \Delta_{t,i} + \bs{b}_y,
\label{eq:y}
\end{equation}
where $\bs{w}_{\bullet, \bullet}, \bs{b}_{\bullet}$ are trainable parameters and $\bs{y}_i \in \mathbb{R}^D$ denotes the vector embedding for the action $e_i$ respectively. In other sections as well, we denote weight and bias as $\bs{w}_{\bullet, \bullet}$ and $\bs{b}_{\bullet, \bullet}$ respectively. 

\xhdr{Self-Attention Layer}
We use a \textit{masked} self-attention layer to embed the past actions to $\bs{s}_k$ and to interpret the influence between the past and the future actions. In detail, we follow the standard attention procedure~\cite{transformer} and first add a trainable positional encoding, $\bs{p}_i$, to the action embedding, \ie, $\bs{y}_i \leftarrow \bs{y}_i + \bs{p}_i$. Such trainable encodings are shown to be more scalable and robust for long sequence lengths as compared to those based on a fixed function~\cite{sasrec,tisasrec}. Later, to calculate an attentive aggregation of all actions in the past, we perform three independent linear transformations on the action representation to get the \textit{query}, \textit{key}, and \textit{value} embeddings, \ie, 
\begin{equation}
\bs{q}_i = \bs{W}^Q  \bs{y}_i, \quad \bs{k}_i = \bs{W}^K  \bs{y}_i, \quad \bs{v}_i = \bs{W}^V  \bs{y}_i,
\end{equation}
where, $\bs{q}_{\bullet}, \bs{k}_{\bullet}, \bs{v}_{\bullet}$ denote the query, key, and value vectors respectively. Following standard self-attention model, we represent $\bs{W}^{Q}$, $\bs{W}^{K}$ and $\bs{W}^{V}$ as trainable \textit{Query}, \textit{Key}, and \textit{Value} matrices respectively. Finally, we compute $\bs{s}_k$ conditioned on the history as:
\begin{equation}
\bs{s}_k =  \sum_{i=1}^{k} \frac{\exp\left(\bs{q}_k^{\top} \bs{k}_i /\sqrt{D} \right)}{\sum_{i'=1}^{k}\exp\left( \bs{q}_k^{\top} \bs{k}_{i'} /\sqrt{D} \right)} \bs{v}_i, \label{eq:attn}
\end{equation}
where $D$ denotes the number of hidden dimensions. Here, we compute the attention weights via a soft-max over the interactions between the query and key embeddings of each action in the sequence and perform a weighted sum of the value embeddings.

Now, given the representation $\bs{s}_k$, we use the attention mechanism in Eqn.~\eqref{eq:attn} and apply a feed-forward neural network to incorporate the necessary non-linearity to the model as:
\begin{equation*}
\bs{s}_k \leftarrow \sum_{i=1}^k \big[ \bs{w}_{s, m}\odot\textsc{ReLU}(\bs{s}_i \odot \bs{w}_{s, n} + \bs{b}_{s, n})  + \bs{b}_{s, m} \big],
\end{equation*}
where, $\bs{w}_{s, m}, \bs{b}_{s, m}$ and $\bs{w}_{s, n}, \bs{b}_{s, n}$ are trainable parameters of the outer and inner layer of the point-wise feed-forward layer.  

To support faster convergence and training stability, we employ the following: (i) layer normalization; (ii) stacking multiple self-attention blocks; and (iii) multi-head attention. Since these are standard techniques~\cite{ba2016layer,transformer}, we omit their mathematical descriptions in this paper.

\xhdr{Output Layer}
At every index $k$, \ourm outputs the next action and the most probable goal of the CTAS. We present the prediction procedure for each of them as follows: \\

\noindent \textit{\underline{Action Prediction:}}
We use the output of the self-attention layer, $\bs{s}_k$ to estimate the mark distribution and time density of the next event, \ie, $\mathbb{P}_{\theta}(e_{k+1})$ and $\rho_{\theta}(e_{k+1})$ respectively. Specifically, we model the $\mathbb{P}_{\theta}(\cdot)$ as a softmax over all other marks as:
\begin{equation}
\mathbb{P}_{\theta}(c_{k+1}) = \frac{\exp\left(\bs{w}_{c, s}^{\top} \bs{s}_i + \bs{b}_{c, s} \right)}{\sum_{c'=1}^{|\cm{C}|}\exp\left( \bs{w}_{c', s}^{\top} \bs{s}_i + \bs{b}_{c', s} \right)},
\label{eqn:samplemark}
\end{equation}
where, $\bs{w}_{\bullet, \bullet}$ and $\bs{b}_{\bullet, \bullet}$ are trainable parameters.

In contrast to standard MTPP approaches that rely on an intensity-based model~\cite{rmtpp,nhp,thp,sahp}, we capture the inter-action arrival times via a \textit{temporal} normalizing flow (NF). In detail, we use a \textit{LogNormal} flow to model the temporal density $\rho_{\theta}(\Delta_{t, k+1})$. Moreover, standard flow-based approaches~\cite{intfree,ppflows} utilize a common NF for all events in a sequence, \ie, the arrival times of each event are determined from a single or mixture of flows trained on all sequences. We highlight that such an assumption restricts the ability to model the dynamics of a CTAS, as unlike standard events, an action has three distinguishable characteristics -- (i) every action requires a minimum time for completion; (ii) the time taken by an actor to complete an action would be similar to the times of another actor; and (iii) similar actions require similar times to complete. For example, the time taken to complete the action \cat{add-coffee} would require a certain minimum time of completion and these times would be similar for all actors. Intuitively, the time for completing the action \cat{add-coffee} would be similar to those for the action \cat{add-sugar}. 

To incorporate these features in \ourm, we identify actions with similar completion times and model them via independent temporal flows. Specifically, we cluster all actions $c_i \in \cm{C}$ into $\cm{M}$ non-overlapping clusters based on the \textit{mean} of their times of completion, and for each cluster, we define a trainable embedding $\bs{z}_r \in \mathbb{R}^{D} \, \forall r \in \{1, \cdots, \cm{M}\}$. Later, we sample the start-time of the future action by conditioning our temporal flows on the cluster of the current action as:
\begin{equation}
\widehat{\Delta_{t, k+1}} \sim \textsc{LogNormal}\left(\bs{\mu}_k , \bs{\sigma}^2_k \right),
\label{eqn:sampletime}
\end{equation}
where, $[\bs{\mu}_k ,\bs{\sigma}^2_k]$, denote the mean and variance of the lognormal temporal flow and are calculated via the sequence embedding and the cluster embedding as:
\begin{equation}
\bs{\mu}_k = \sum_{r=1}^{\cm{M}} \cm{R}(e_k, r) \big(\bs{w}_{\mu} \left(\bs{s}_{k}\odot\bs{z}_{c, i} \right) + \bs{b}_{\mu}\big),
\end{equation}
\begin{equation}
\bs{\sigma}^2_k  = \sum_{r=1}^{\cm{M}} \cm{R}(e_k, r)  \big( \bs{w}_{\sigma} \left(\bs{s}_{k}\odot\bs{z}_{c, i} \right) + \bs{b}_{\sigma}\big),
\end{equation}
where $\bs{w}_{\bullet}, \bs{b}_{\bullet}$ are trainable parameters, $\cm{R}(e_k, r)$ is an indicator function that determines if event $e_k$ belongs to the cluster $r$ and $\bs{z}_r$ denotes the corresponding cluster embedding. Such a cluster-based formulation facilitates the ability of the model to assign similar completion times for events in the same cluster. To calculate the time of the next action, we add the sampled time difference to the time of the previous action $e_k$, \ie,
\begin{equation}
\widehat{t_{k+1}} = t_k + \widehat{\Delta_{t, k+1}},
\end{equation}
where, $\widehat{t_{k+1}}$ denotes the predicted time for the action $e_{k+1}$. \\

\noindent \textit{\underline{Goal Detection:}}
In contrast to other MTPP approaches~\cite{rmtpp,nhp,thp,sahp,intfree}, an important feature of \ourm is identifying the goal of a sequence, \ie, a hierarchy on top of the actions in a sequence, based on the past sequence dynamics. To determine the goal of a CTAS, we utilize the history embedding $\bs{s}_k$ as it encodes the inter-action relationships of all actions in the past. Specifically, we use a non-linear transformation via a feed-forward network, denoted as $\Phi(\cdot)$ over $\bs{s}_k$ and apply a softmax over all possible goals.
\begin{equation}
\Phi(\bs{s}_k) = \textsc{ReLU} (\bs{w}_{\Phi, s} \bs{s}_k + \bs{b}_{\Phi, s}),
\end{equation}
where, $\bs{w}_{\bullet, \bullet}, \bs{b}_{\bullet, \bullet}$ are trainable parameters. We sample the most probable goal as in Eqn.~\eqref{eq:goal}. We highlight that we predict the CTAS goal at each interval, though a CTAS has only one goal. This facilitates \textit{early} goal detection in comparison to detecting the goal after traversing the entire CTAS. More details are given in Section~\ref{sec:early} and Section~\ref{sec:optimization}.

\subsection{Early Goal Detection and Action Hierarchy}\label{sec:early}
Here, we highlight the two salient features of \ourm -- early goal detection and modeling the goal-action hierarchy.

\xhdr{Early Goal Detection}
Early detection of sequence goals has many applications ranging from robotics to vision~\cite{earlyiccv,earlyijcv}. To facilitate early detection of the goal of a CTAS in \ourm, we devise a ranking loss that forces the model to predict a \textit{non-decreasing} detection score for the correct goal category. Specifically, the detection score of the correct goal at the $k$-th index of the sequence, denoted by $p_k(g| \bs{s}_k, \Phi)$, must be more than the scores assigned the correct goal in the past. Formally, we define the ranking loss as:
\begin{equation}
\cm{L}_{k, g} = \max \big(0, p^*_k(g) - p_k(g| \bs{s}_k, \Phi)\big),
\label{eqn:margin}
\end{equation}
where $p^*_k(g)$ denotes the maximum probability score given to the correct goal in all past predictions.
\begin{equation}
p^*_k(g) = \max_{j \in \{1, k-1\}} p_j(g| \bs{s}_j, \Phi),
\label{eqn:pastmax}
\end{equation}
where $p_k(g)$ denotes the probability score for the correct goal at index $k$. Intuitively, the ranking loss $\cm{L}_{k,g}$ would penalize the model for predicting a smaller detection score for the correct CTAS goal than any previous detection score for the same goal.

\xhdr{Action Hierarchy}
Standard MTPP approaches assume the category of marks as independent discrete variables, \ie, the probability of an upcoming mark is calculated independently~\cite{rmtpp,nhp,sahp,thp}. Such an assumption restricts the predictive ability while modeling CTAS, as in the latter case, there exists a hierarchy between goals and actions that lead to the specific goal. Specifically, actions that lead to a common goal may have similar dynamics and it is also essential to model the relationships between the actions of different CTAS with a common goal. We incorporate this hierarchy in \ourm along with our next action prediction via an action-based ranking loss. In detail, we devise a loss function similar to Eqn.~\eqref{eqn:margin} where we restrict the model to assign non-decreasing probabilities to all actions leading to the goal of CTAS under scrutiny. 
\begin{equation}
\cm{L}_{k, c} = \sum_{c' \in \cm{C}^*_g} \max \big(0, p^*_k(c') - p_k(c'| \bs{s}_k)\big),
\label{eqn:cat_margin}
\end{equation}
where $\cm{C}^*_g, p_k(c'| \bs{s}_k)$ denote a set of all actions in CTAS with the goal $g$ and the probability score for the action $c' \in \cm{C}^*_g$ at index $k$ respectively. Here, $p^*_k(c')$ denotes the maximum probability score given to action $c'$ in all past predictions and is calculated similar to Eqn.~\eqref{eqn:pastmax}. We regard  $\cm{L}_{k, g}$ and $\cm{L}_{k, c}$ as \textit{margin} losses, as they aim to increase the difference between two prediction probabilities.

\subsection{Optimization}\label{sec:optimization}
We optimize the trainable parameters in \ourm, \ie, the weight and bias tensors ($\bs{w}_{\bullet, \bullet}$ and $\bs{b}_{\bullet, \bullet}$) for our MTPP $p_{\theta} (\cdot)$, using two channels of training consisting of action and goal prediction. Specifically, to optimize the ability of \ourm for predicting the next action, we maximize the joint likelihood for the next action and the lognormal density distribution of the temporal flows.
\begin{equation}
\mathscr{L} = \sum_{k = 1}^{|\cm{S}|} \log \big( \mathbb{P}_{\theta}(c_{k+1}|\bs{s}_k) \cdot \rho_{\theta} (\Delta_{t, k+1}| \bs{s}_k) \big ),
\label{eqn:likelihood}
\end{equation}
where $\mathscr{L}$ denotes the joint likelihood, which we represent as the sum of the likelihoods for all CTAS. In addition to action prediction, we optimize the \ourm parameters for \textit{early} goal detection via a temporally weighted cross entropy (CE) loss over all sequence goals. Specifically, we follow a popular reinforcement recipe of using a time-varying \textit{discount} factor over the prediction loss as:
\begin{equation}
\cm{L}_g = \sum_{k = 1}^{|\cm{S}|} \gamma^k \cdot \cm{L}_{\textsc{CE}} \big(p_k(g| \bs{s}_k)\big),
\label{eqn:discount}
\end{equation}
where $\gamma \in [0,1], \cm{L}_{\textsc{CE}} \big(p_k(g| \bs{s}_k)\big)$ denote the decaying factor and a standard softmax cross-entropy loss respectively. Such a recipe is used exhaustively for faster convergence of reinforcement learning models~\cite{rl_book,discount}. Here, the discount factor penalizes the model for taking longer times for detecting the CTAS goal by decreasing the gradient updates to the loss.

\xhdr{Margin Loss}
In addition, we minimize the margin losses given in Section~\ref{sec:early} with the current optimization procedure. Specifically, we minimize the following loss:
\begin{equation}
\cm{L}_m = \sum_{k=1}^{|\cm{S}|} \cm{L}_{k, g} + \cm{L}_{k, c},
\end{equation}
where $\cm{L}_{k, g}$ and $\cm{L}_{k, c}$ are margin losses defined in Eqn.~\eqref{eqn:margin} and Eqn.~\eqref{eqn:cat_margin} respectively. We learn the parameters of \ourm using an Adam~\cite{adam} optimizer for both likelihood and prediction losses.

\subsection{Sequence Generation}\label{sec:generation}
A crucial contribution of this paper via \ourm is an end-to-end generation of action sequences. Specifically, given the CTAS goal as input, we can generate the most probable sequence of actions that may lead to that specific goal. Such a feature has a range of applications from sports analytics~\cite{mehrasa2017learning}, forecasting~\cite{prathamesh}, identifying the duration of an activity~\cite{avae}, \etc\ A standard approach for training a sequence generator is to sample future actions in a sequence and then compare with the true actions~\cite{timegan}. However, such a procedure has multiple drawbacks as it is susceptible to noises during training and deteriorates the scalability of the model. Moreover, we highlight that such sampling-based training cannot be applied to a self-attention-based model as it requires a fixed-sized sequence as input~\cite{transformer}. Therefore, we resort to a two-step generation procedure that is defined below:
\begin{enumerate}
\item \textbf{Pre-Training:} The first step requires training all \ourm parameters for action prediction and goal detection. This step is necessary to model the relationships between actions and goals and we represent the set of optimized parameters as $\theta^*$ and the corresponding MTPP as $p_{\theta^*}(\cdot)$ respectively.
\item \textbf{Iterative Sampling:} We iteratively sample events and update parameters via our trained MTPP till the model predicts the correct goal for the CTAS or we encounter an \texttt{<EOS>} or end-of-sequence action. Specifically, using $p_{\theta^*}(\cdot)$ and the first \textit{real} action ($e_1$) as input, we calculate the detection score for the correct goal, \ie, $p_1(g| \bs{s}_k)$ and while its value is highest among all probable goals, we sample the mark and time of next action using Eqn.~\eqref{eqn:samplemark} and Eqn.~\eqref{eqn:sampletime} respectively.
\end{enumerate}
Such a generation procedure harnesses the fast sampling of temporal normalizing flows and simultaneously is conditioned on the action and goal relationships. A detailed pseudo-code of sequence generation procedure used in \ourm is given in Algorithm~\ref{axoalgo}.

\begin{figure}[t!]
\centering
\begin{minipage}{.7\linewidth}
\begin{algorithm}[H]
\small
\textbf{Input:} $g$: Goal of CTAS, $e_1$: First Action, $p_{\theta^*}(\cdot)$: Trained MTPP\\
\textbf{Output:} $\widehat{S}$: Generated CTAS
$\cm{S}_1 \leftarrow e_1$ \\
$k = 1$\\ 
  \While {$k < \mathtt{max\_len}$}
  {
    Sample the mark of next action: $\widehat{c_{k+1}} \sim \mathbb{P}_{\theta^*}(\bs{s}_k)$\\
    Sample the time of next action: $\widehat{t_{k+1}} \sim \rho_{\theta^*}(\bs{s}_k)$\\
    Add to CTAS: $\cm{S}_{k+1} \leftarrow \cm{S}_{k} + e_{k+1}$\\
    Update the MTPP parameters $\bs{s}_{k+1} \leftarrow p(\bs{s}_{k}, e_{k+1})$\\
    Calculate most probable goal: $\widehat{g}_k = \max_{\forall g'} \big(p_k(g'| \bs{s}_k)\big)$\\
    \uIf{$\widehat{g_i} != g \, \mathrm{or} \, \widehat{c_{k+1}} == \mathtt{<EOS>}$}
    {
    Add EOS mark: $\widehat{\cm{S}} \leftarrow \cm{S}_{k+1} + \mathtt{<EOS>}$\\
    Exit the sampling procedure: $\textsc{Break}$\\
    }
    Increment iteration: $k \leftarrow k + 1$\\
    
  }\label{endfor}
  Return generated CTAS: return $\widehat{\cm{S}}$\\
\caption{Sequence Generation with \ourm}\label{axoalgo}
\end{algorithm}
\end{minipage}
\vspace{-0.3cm}
\end{figure}

\section{\ourp: Modeling Variations in Order of Actions}
In standard time-series datasets, we assume that the events recorded are \textit{causal} in nature, \ie, the future events have been influenced by the events in the past. However, in a CTAS, the actor can determine the order in which they want to perform a given set of actions~\cite{felsen_thesis, felsen_iccv, felsen_eccv}. For \eg, while cooking an omelet, an actor might add salt before adding pepper or vice-versa. Moreover, the order of actions can vary significantly between different actors. Thus, assuming that the actions have an inherent causality can deteriorate the ability of a model to learn the dynamics of CTAS, and idealistically, the model performance should be invariant to these perturbations.

This problem of \textit{permutation-invariant} data modeling is present in a wide range of applications, such as 3D shape recognition and few-shot image classification. Recent research has shown that deep-learning approaches can be made permutation invariant by utilizing \textit{set}-based embedding techniques~\cite{deepsets, settrans}. Since, sets are inherently invariant of the order of objects within them, incorporating the past actions as a set can make the model resistant to variations in the action order. We formally define a permutation invariant network as follows:

\begin{definition}[Permutation Invariant Sequential Network]
\textit{A set of objects has a permutation invariant property, \ie, the resultant output for a given sequence is the same regardless of the order of occurrence of events in the sequence. For \eg, a permutation invariant model is a network that has a pooling function over the embeddings obtained from the events in a sequence. More formally, any $\cm{F}(\cdot)$ is considered permutation invariant if it has the following formulation:}
\begin{equation}
\cm{F}(\cm{S}_k) = \cm{F} \left (\{e_1, \cdots , e_k\} \right ) = \rho \big(\textsc{Pool} ( \{ \phi(x_1), \cdots , \phi(x_n)\}) \big ),
\end{equation}
\textit{where $\textsc{Pool}(\cdot)$ is a summation operator, and $\rho(\cdot)$ and $\phi(\cdot)$ are continuous functions, which can be approximated by MLPs.}
\label{def:pisn}
\end{definition}

\subsection{Permutation Invariant Modeling of CTAS}
To provide better and more personalized action recommendations, we introduce a permutation invariant CTAS modeling ability in \ourm. Here, we highlight the changes required in the neural formulation to make it more robust to variations in the CTAS order. Since these capabilities are incorporated in addition to the sequential MTPP framework, we name it \ourp. In detail, \ourp has the following enhancements:

\xhdr{Set Embedding Layer}
In Eqn.~\eqref{eq:y}, we get a vector embedding, $\bs{y}_{k}$, for an action, $e_{k}$, by capturing its time, action-type, and the latest time-interval. In \ourp, we use the embeddings for all the past actions to calculate the \textit{set} embedding of history, \ie, actions till index $k$, as follows: 
\begin{equation}
    \bs{x}_k = \sum_{i=1}^{k} \textsc{ReLU} \left ( \Omega(\bs{w}_{x} \bs{y}_i + \bs{b}_{x}) \right ),
    \label{eqn:set_emb}
\end{equation}
where, $\bs{x}_k$ denotes the embedding of the set of actions till index $k$, $\Omega(\cdot)$ denotes the transformation via a fully-connected MLP layer, and $\bs{w}_{\bullet}$ and $\bs{b}_{\bullet}$ are trainable parameters. This formulation references with Definition~\ref{def:pisn} as follows: (i) the layer $\Omega(\cdot)$ followed by a $\textsc{ReLU}(\cdot)$ is continuous; (ii) the formulation using $\bs{w}_{x}$ and $\bs{b}_{x}$ is also continuous; and (iii) the $\textsc{Pool}(\cdot)$ function is replaced by the sum operator.

\xhdr{Next Action Prediction}
To predict the category of the incoming action in the CTAS, we use the output of the self-attention layer, $\bs{s}_k$, and the set-embedding, $\bs{x}_k$ to estimate the mark distribution of actions. In detail, we use the following formulation:
\begin{equation}
\mathbb{P}_{\theta, \bs{x}_k}(c_{k+1}) = \frac{\exp\left(\bs{w}_{c, x}^{\top} (\bs{s}_i +  \alpha_{c} \bs{z}_{k})  + \bs{b}_{c, x} \right)}{\sum_{c'=1}^{|\cm{C}|}\exp \big( \bs{w}_{c', x}^{\top} (\bs{s}_i + \alpha{c}\bs{z}_{k}) + \bs{b}_{c', x} \big)},
\label{eqn:new_samplemark}
\end{equation}
where, $\mathbb{P}_{\theta, \bs{x}_k}(\cdot)$ refers to the distribution of actions-types conditioned on the set embedding, $\alpha_{c}$ denotes weighing hyper-parameter for $\bs{x}_k$, and $\bs{w}_{\bullet, \bullet}$ and $\bs{b}_{\bullet, \bullet}$ are trainable parameters. Note that we keep the same value of $\alpha_{c}$ for all the action categories. We use a similar method for time prediction, \ie, combining the self-attention output and the set embedding for the log-normal flow. 

\xhdr{CTAS Goal Detection}
Similar to action prediction, we combine the set embedding with the output of the self-attention layer to get a combined history embedding. In detail, we detect the goal of the CTAS as follows:
\begin{equation}
\Phi(\bs{s}_k, \bs{x}_k) = \textsc{ReLU} \left ( \bs{w}_{\Phi, s} (\bs{s}_i +  \alpha_{g} \bs{z}_{k}) + \bs{b}_{\Phi, s} \right ),
\label{eq:new_g}
\end{equation}
where, $\bs{w}_{\bullet, \bullet}, \bs{b}_{\bullet, \bullet}$ are trainable parameters and $\alpha_{g}$ denotes weighing hyper-parameter for $\bs{x}_k$. As in \ourm, we sample the most probable goal as in Eqn.~\eqref{eq:goal}.

\xhdr{CTAS Generation}
The CTAS generation procedure of \ourp is similar to the description given in Section~\ref{sec:generation}, \ie, we predict the goal of the CTAS at each generated action. However, here we also use the set embedding to predict the goal of the generated CTAS. In detail, we use the following steps:
\begin{enumerate}
    \item[(1)] \textbf{Embed the Sampled Actions:} For each action-type and time ($\widehat{c_{\bullet}}$ and $\widehat{t_{\bullet}}$ respectively) sampled by the generative procedure, we calculate the action embedding using Eqn.~\eqref{eq:y}, and the corresponding set embedding using Eqn.~\eqref{eqn:set_emb}. 
    
    \item[(2)] \textbf{Most Probable Goal:} Later, we use the procedure described in Eqn.~\eqref{eq:new_g} to sample the most likely goal of the generated CTAS. We compare the predicted goal with the provided \textit{true} CTAS goal to stop or continue the sampling procedure. 
\end{enumerate}
We highlight that the set embedding keeps the CTAS generation process in better sync with the history, \ie, it allows the model to prevent an action that has occurred in the past from being sampled in the immediate future.

\section{Experiments}\label{sec:exp}
In this section, we present the experimental setup and the empirical results to validate the efficacy of \ourm. Through our experiments, we aim to answer the following research questions: 
\begin{itemize}
\item[\textbf{RQ1}] What is the action-mark and time prediction performance of \ourm in comparison to the state-of-the-art baselines?
\item[\textbf{RQ2}] How accurately and quickly can \ourm identify the goal of an activity sequence?
\item[\textbf{RQ3}] How effectively can \ourm generate an action sequence?
\item[\textbf{RQ4}] How does the action prediction performance  of \ourm vary with different hyper-parameter values?
\end{itemize}

\subsection{Datasets}
To evaluate \ourm, we need time-stamped action sequences and their goals. Therefore, we derive CTAS from three activity modeling datasets sourced from different real-world applications -- cooking, sports, and collective activity. The datasets vary significantly in terms of origin, sparsity, and sequence lengths. We highlight the details of each of these datasets below: 
\begin{itemize}
\item \textbf{\bfast~\cite{breakfast}.} This dataset contains CTAS derived from 1712 videos of different people preparing breakfast. The actions in a CTAS and sequence goals can be classified into 48 and 9 classes respectively. These actions are performed by 52 different individuals in 18 different kitchens.

\item \textbf{\mult~\cite{thumos}.} A sports activity dataset that is designed for action recognition in videos.  We derive the CTAS using 400 videos of individuals involved in different sports such as discus throw, baseball, \etc\ The actions and goals can be classified into 65 and 9 classes respectively, and on average, there are 10.5 action class labels per video.

\item \textbf{\act~\cite{activitynet}.} This dataset comprises activity categories collected from 591 YouTube videos with a total of 49 action labels and 14 goals. 

\item \textbf{\actnew.} In \act, many of the videos are shot by amateurs in many uncontrolled environments, the variances within the CTAS of the same goal are often large, and the lengths of CTAS vary and are often long and complex. Thus, we create a separate dataset that consists of sparse action sequences, \ie, a sequence with \textit{only} a few actions. Through this dataset, we aim to highlight the data needs for every model.
\end{itemize}

\subsection{Baselines}
We compare the performance of \ourm with the following state-of-the-art methods:
\begin{asparaitem}[]

\item \textbf{HP~\cite{hawkes}}: Hawkes process or self-exciting multivariate point process model with an exponential kernel. In this model, the intensity of the next event of the same type is increased by past events.

\item \textbf{RMTPP~\cite{rmtpp}}: The underlying model of RMTPP is a two-step procedure that embeds the event sequence using a recurrent neural network (RNN) and then derives the formulation of CIF using this embedding. Specifically, given the sequence, an RNN determines its vector representation. Later, RMTPP uses this representation over an \textit{exponential} function to formulate the CIF. 

\item \textbf{NHP~\cite{nhp}}: NHP modified the LSTM architecture to model the continuous time of events in a sequence. Later, it uses the embedding from the LSTM to determine the CIF using a \textit{soft-plus} function. NHP is more expressive; however, it does not have a closed form for the likelihood.

\item \textbf{AVAE~\cite{avae}}: A variational auto-encoder-based MTPP framework designed specifically for activities in a sequence. AVAE is the \textit{most} relevant approach to our paper as it was designed to handle activity videos. 

\item \textbf{SAHP~\cite{sahp}}: A self-attention model to learn the temporal dynamics using an aggregation of historical events. The intensity-function is formulated by an exponential decay. 

\item \textbf{THP~\cite{thp}}: THP combines the transformer architecture to formulate a point process. In detail, these architectures obtain the embedding of the sequence using a transformer and then formulate the CIF using the obtained embeddings. 
\end{asparaitem}
We omit comparison with other continuous-time models~\cite{fullyneural,intfree,wgantpp,xiao,hawkes} as they have already been outperformed by these approaches. 

\xhdr{Baseline Implementations} For the baselines RMTPP\footnote{https://github.com/musically-ut/tf\_rmtpp.}, NHP\footnote{https://github.com/HMEIatJHU/neurawkes.}, SAHP\footnote{https://github.com/QiangAIResearcher/sahp\_repo}, and THP\footnote{https://github.com/SimiaoZuo/Transformer-Hawkes-Process.} we use the official implementations made public by the authors. To generate a sequence of events of a specified length using HP, we generate $|N|$ sequences based on the learned weights from the training set. These sequences are denoted as $\mathcal{S} = \{s_1, s_2, \cdots s_N\}$, each with a maximum sequence length. For evaluation purposes, we consider the first $l_i$ set of events from each sequence $i$. For RMTPP, we set hidden dimension and BPTT is selected among \{32, 64\} and \{20, 50\} respectively. For THP and SAHP, we set the number of attention heads as 2, hidden key-matrix, and value-matrix dimensions are selected among \{32, 64\}. If applicable, for each model we use a dropout of 0.1. All other parameter values are the ones recommended by the authors of the corresponding models. 

\subsection{Evaluation Criteria}
Given the dataset $\cm{D}$ of $N$ action sequences, we split them into training and test sets based on the goal of the sequence. Specifically, for each goal $g \in \cm{G}$, we consider 80\%  of the sequences as the training set and the other last 20\% as the test set. We evaluate \ourm and all baselines on the test set in terms of (i) mean absolute error (MAE) of predicted times of action, and (ii) action prediction accuracy (APA) described as:
\begin{equation}
\mathrm{MAE} = \frac{1}{|\cm{S}|}\sum_{e_i\in \cm{S}}[|t_i-\widehat{t}_i|], \quad \mathrm{APA} = \frac{1}{|\cm{S}|}\sum_{e_i\in \cm{S}} \#(c_i=\widehat{c}_i),
\end{equation}
where, $\widehat{t_i}$ and $\widehat{c_i}$ are the predicted time and type the $i$-th action in test set. Moreover, we follow a similar protocol to evaluate the sequence generation ability of \ourm and other models. We use a similar procedure to calculate the goal prediction performance, \ie, GPA, of all the models. In detail, we use any model to predict the end goal of the sequence and then average the results across all the CTAS in the dataset. We calculate confidence intervals across 5 independent runs.

\subsection{Experimental Setup}
The implementations for \ourm are available at: \texttt{https://github.com/data-iitd/proactive}.

\xhdr{System Configuration}
All our experiments were done on a server running Ubuntu 16.04. CPU: Intel(R) Xeon(R) Gold 5118 CPU @ 2.30GHz , RAM: 125GB and GPU: NVIDIA V100 32GB. 

\xhdr{Parameter Settings}
For our experiments involving \ourm and \ourp, we set $D$=16, $\cm{M}$=8, $\gamma$=0.9 and weigh the margin loss $\cm{L}_m$ by 0.1. In addition, we set a $l_2$ regularizer over the parameters with a coefficient value of 0.001.

\begin{table}[t!]
\small
\caption{Performance of all the methods in terms of action prediction accuracy (APA). Bold (underline) fonts indicate the best performer (baseline). Results marked \textsuperscript{$\dagger$} are statistically significant (\ie, two-sided Fisher's test with $p \le 0.1$) over the best baseline. \vspace{-0.2cm}}
\centering
\begin{tabular}{l|cccc}
\toprule
\textbf{Dataset} & \multicolumn{4}{c}{\textbf{Action Prediction Accuracy (APA)}} \\ \hline 
 & \bfast & \mult & \act & \actnew \\ \hline \hline
HP~\cite{hawkes} & 0.503$\pm$0.013 & 0.257$\pm$0.014 & 0.651$\pm$0.023 & 0.431$\pm$0.021\\
NHP~\cite{nhp} & 0.528$\pm$0.024 & 0.272$\pm$0.019 & 0.684$\pm$0.034 & 0.464$\pm$0.029\\
AVAE~\cite{avae} & 0.533$\pm$0.028 & 0.279$\pm$0.022 & 0.678$\pm$0.036 & 0.452$\pm$0.030\\
RMTPP~\cite{rmtpp} & 0.542$\pm$0.022 & 0.274$\pm$0.017 & 0.683$\pm$0.034 & 0.470$\pm$0.029\\
SAHP~\cite{sahp} & 0.547$\pm$0.031 & 0.287$\pm$0.023 & 0.688$\pm$0.042 & 0.467$\pm$0.036\\
THP~\cite{thp} & \underline{0.559$\pm$0.028} & \underline{0.305$\pm$0.018} & \underline{0.693$\pm$0.038} & \underline{0.489$\pm$0.033}\\
\hline
\ourm & 0.583$\pm$0.027 & 0.316$\pm$0.019 & 0.728$\pm$0.037 & 0.521$\pm$0.034\\
\ourp & \textbf{0.601$\pm$0.029}\textsuperscript{$\dagger$} & \textbf{0.334$\pm$0.019}\textsuperscript{$\dagger$} & \textbf{0.746$\pm$0.039}\textsuperscript{$\dagger$} & \textbf{0.536$\pm$0.035}\textsuperscript{$\dagger$}\\
\bottomrule
\end{tabular}
\label{tab:apa}
\vspace{0.3cm}
\caption{Performance of all the methods in terms of mean absolute error (MAE). Similar to Table~\ref{tab:apa}, bold (underline) fonts indicate the best performer (baseline). Results marked \textsuperscript{$\dagger$} are statistically significant (\ie, two-sided Fisher's test with $p \le 0.1$) over the best baseline. \vspace{-0.2cm}}
\begin{tabular}{l|cccc}
\toprule
\textbf{Dataset} & \multicolumn{4}{c}{\textbf{Mean Absolute Error (MAE)}} \\ \hline 
 & \bfast & \mult & \act & \actnew \\ \hline \hline
HP~\cite{hawkes} & 0.422$\pm$0.026 & 0.018$\pm$0.002 & 0.784$\pm$0.043 & 1.184$\pm$0.047 \\
NHP~\cite{nhp} & 0.411$\pm$0.019 & \underline{0.017$\pm$0.002} & 0.796$\pm$0.045 & 1.168$\pm$0.041 \\
AVAE~\cite{avae} & 0.417$\pm$0.021 & 0.018$\pm$0.002 & 0.803$\pm$0.049 & 1.163$\pm$0.043 \\
RMTPP~\cite{rmtpp} & \underline{0.403$\pm$0.018} & \underline{0.017$\pm$0.002} & \underline{0.791$\pm$0.046} & \underline{1.157$\pm$0.044} \\
SAHP~\cite{sahp} & 0.425$\pm$0.031 & 0.019$\pm$0.003 & 0.820$\pm$0.072 & 1.211$\pm$0.071 \\
THP~\cite{thp} & 0.413$\pm$0.023 & 0.019$\pm$0.002 & 0.806$\pm$0.061 & 1.174$\pm$0.058 \\
\hline
\ourm & 0.364$\pm$0.028 & \textbf{0.013$\pm$0.002}\textsuperscript{$\dagger$} & 0.742$\pm$0.059 & 1.083$\pm$0.053 \\
\ourp & \textbf{0.362$\pm$0.029}\textsuperscript{$\dagger$} & \textbf{0.013$\pm$0.002}\textsuperscript{$\dagger$} & \textbf{0.739$\pm$0.057}\textsuperscript{$\dagger$} & \textbf{1.071$\pm$0.052}\textsuperscript{$\dagger$} \\
\bottomrule
\end{tabular}
\label{tab:mae}
\end{table}

\subsection{Action Prediction Performance (RQ1)}
We evaluate the action prediction performance of \ourm and the baselines using action prediction accuracy (APA) and mean absolute error (MAE). The results across all our datasets are reported in Table~\ref{tab:apa} and Table~\ref{tab:mae} respectively. From the results, we make the following observations:
\begin{asparaitem}[$\bullet$]
\item The \ourm model consistently yields a superior prediction performance over all the baseline models. In particular, it improves over the strongest baselines by 8-27\% for time prediction and by 2-7\% for action prediction. These results signify the drawbacks of using standard sequence approaches for modeling a temporal action sequence. 

\item Modeling the variations in the order of actions via \ourp leads to an improvement over the standard \ourm model, However, this improvement is significant only for \bfast dataset. Thus, we highlight that the other datasets have limited variations in their sequentiality, and the order in which actions are performed does not vary significantly across different actors. 

\item RMTPP~\cite{rmtpp} is the second-best performer in terms of MAE of time prediction in almost all the datasets. We also note that for \act\ dataset, THP~\cite{thp} outperforms RMTPP for action category prediction. However, \ourm and \ourp still significantly outperform these models across all metrics.

\item Variations of MTPP methods that leverage self-attention mechanisms improve the modeling of action distributions. In particular, three notable approaches, namely THP, SAHP, \ourm, and \ourp demonstrate enhanced performance in terms of category prediction. This result reinforces the need to design MTPP frameworks that can handle long-term dependencies within a CTAS or time series in general.

\item AVAE~\cite{avae} is a sequence model specifically designed for modeling activity sequences. While AVAE has its strengths, recent advancements in neural methods incorporating complex structures, such as self-attention or normalizing flows, have demonstrated superior performance compared to AVAE in various tasks.
\end{asparaitem}
\noindent To sum up, our empirical analysis suggests that \ourm can better model the underlying dynamics of a CTAS as compared to all other baseline models.

\begin{figure}[t]
\centering
\hfill
\begin{subfigure}[b]{0.30\columnwidth}
\centering
\includegraphics[height=3cm]{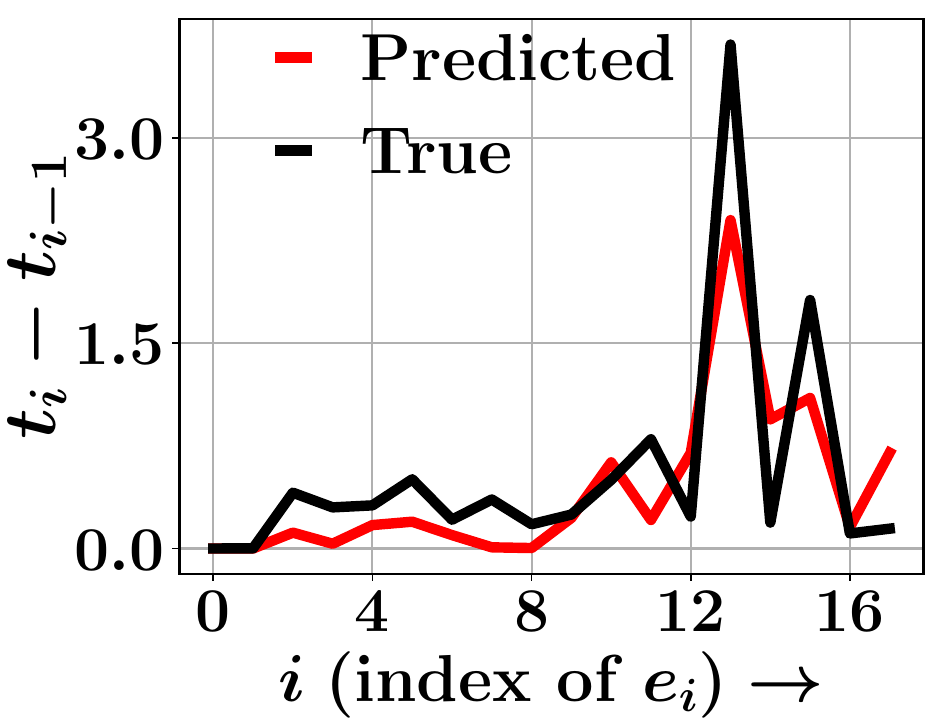}
\caption{\bfast}
\end{subfigure}
\hfill
\begin{subfigure}[b]{0.30\columnwidth}
\centering
\includegraphics[height=3cm]{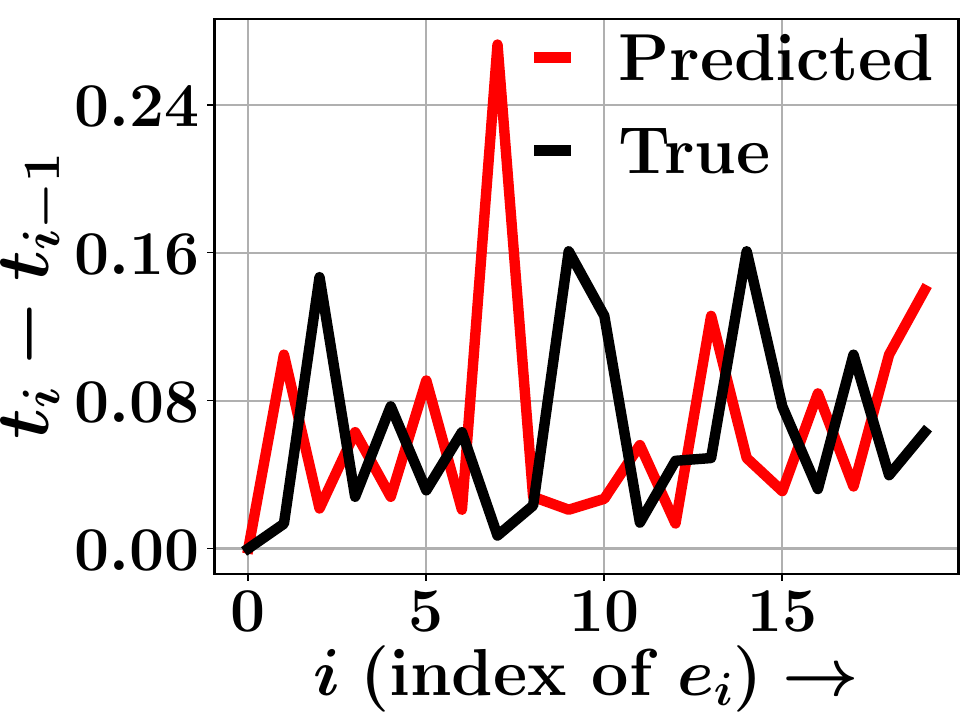}
\caption{\mult}
\end{subfigure}
\hfill
\begin{subfigure}[b]{0.3\columnwidth}
\centering
\includegraphics[height=3cm]{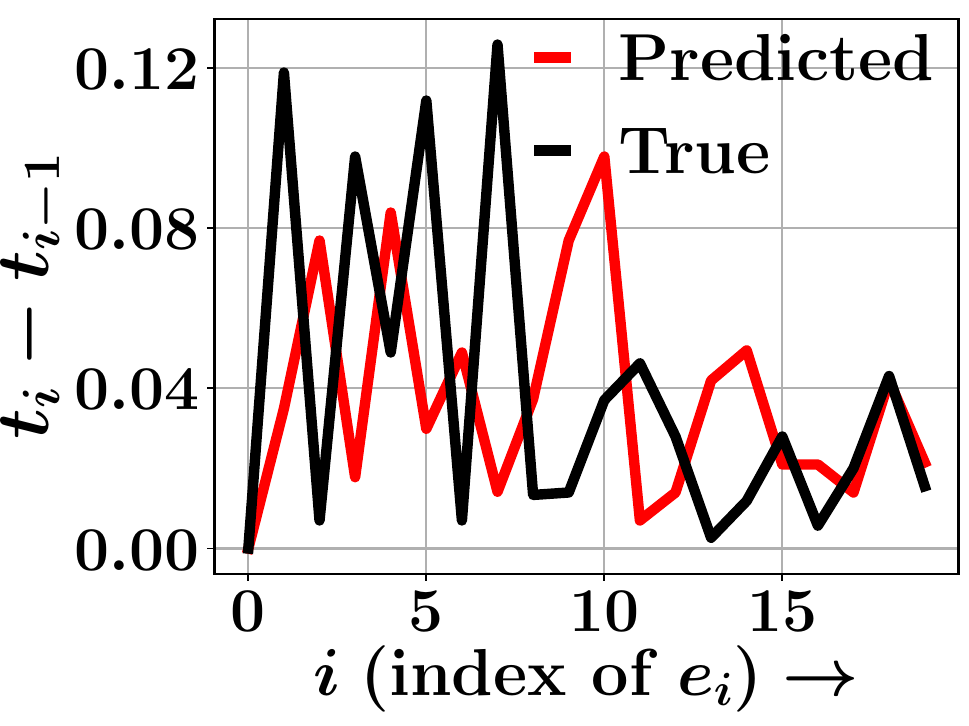}
\caption{\act}
\end{subfigure}
\hfill
\vspace{-0.3cm}
\caption{\label{fig:qualitative} Real life \textit{true} and \textit{predicted} inter-arrival times $\Delta_{t,k}$ of different events $e_k$ for \bfast, \mult, and \act\ datasets. The results show that the true arrival times match with the times predicted by \ourm. \vspace{-0.3cm}}
\end{figure}

\xhdr{Qualitative Assessment}
We also perform a qualitative analysis to highlight the ability of \ourm for modeling the inter-arrival times for action prediction. Specifically, we plot the actual inter-action time differences and the time differences predicted by \ourm in Figure~\ref{fig:qualitative} for \bfast, \mult, and \act\ datasets. From the results, we note that the predicted inter-arrival times closely match with the true inter-arrival times, and \ourm is even able to capture large time differences (peaks). For brevity, we omitted the results for the \actnew\ dataset.

\begin{table}[t!]
\small
\caption{Performance of all variants of \ourm in terms of action prediction accuracy (APA). \vspace{-0.2cm}}
\centering
\begin{tabular}{l|cccc}
\toprule
\textbf{Dataset} & \multicolumn{4}{c}{\textbf{Action Prediction Accuracy (APA)}} \\ \hline 
 & \bfast & \mult & \act & \actnew \\ \hline \hline
\ours-c & 0.561$\pm$0.027 & 0.297$\pm$0.020 & 0.698$\pm$0.038 & 0.495$\pm$0.036\\
\ours-t & 0.579$\pm$0.025 & 0.306$\pm$0.018 & 0.722$\pm$0.035 & 0.513$\pm$0.032 \\
\ourm & \textbf{0.583$\pm$0.027} & \textbf{0.316$\pm$0.019} & \textbf{0.728$\pm$0.037} & \textbf{0.521$\pm$0.034}\\
\bottomrule
\end{tabular}
\label{tab:abs_apa}
\vspace{0.3cm}

\caption{Performance of all variants of \ourm in terms of mean absolute error (MAE). \vspace{-0.2cm}}
\begin{tabular}{l|cccc}
\toprule
\textbf{Dataset} & \multicolumn{4}{c}{\textbf{Mean Absolute Error (MAE)}} \\ \hline 
 & \bfast & \mult & \act & \actnew \\ \hline \hline
\ours-c & 0.415$\pm$0.027 & 0.015$\pm$0.002 & 0.774$\pm$0.054 & 1.169$\pm$0.051\\
\ours-t & 0.407$\pm$0.025 & 0.015$\pm$0.002 & 0.783$\pm$0.058 & 1.197$\pm$0.057\\
\ourm & \textbf{0.364$\pm$0.028} & \textbf{0.013$\pm$0.002} & \textbf{0.742$\pm$0.059} & \textbf{1.083$\pm$0.053} \\
\bottomrule
\end{tabular}
\label{tab:abs_mae}
\end{table}

\subsection{Ablation Study}
We conduct an ablation study for two key contributions in \ourm, \ie, goal-action hierarchy loss and cluster-based flows. In detail, we include two variants of \ourm -- (i) \ours-c, which represents our model without the goal-action hierarchy loss and cluster-based flows, and (ii) \ours-t, which represents our model without cluster-based flows. We report the results in terms of APA and MAE in Table~\ref{tab:abs_apa} and Table~\ref{tab:abs_mae} respectively. The results show that \ours-c performs better than \ours-t, and thus reinforces the need to capture the relationship between actions and the CTAS goal. These results further signify the need to use a permutation invariant approach via \ourp, that more effectively models the action-goal hierarchy. However, the performance gain \ourm has over both, \ours-c and \ours-t,  highlights the need for modeling action hierarchy as well as the cluster-based flows. 

\begin{figure}[t]
\centering
\hfill
\begin{subfigure}[b]{0.30\columnwidth}
\centering
\includegraphics[height=2.7cm]{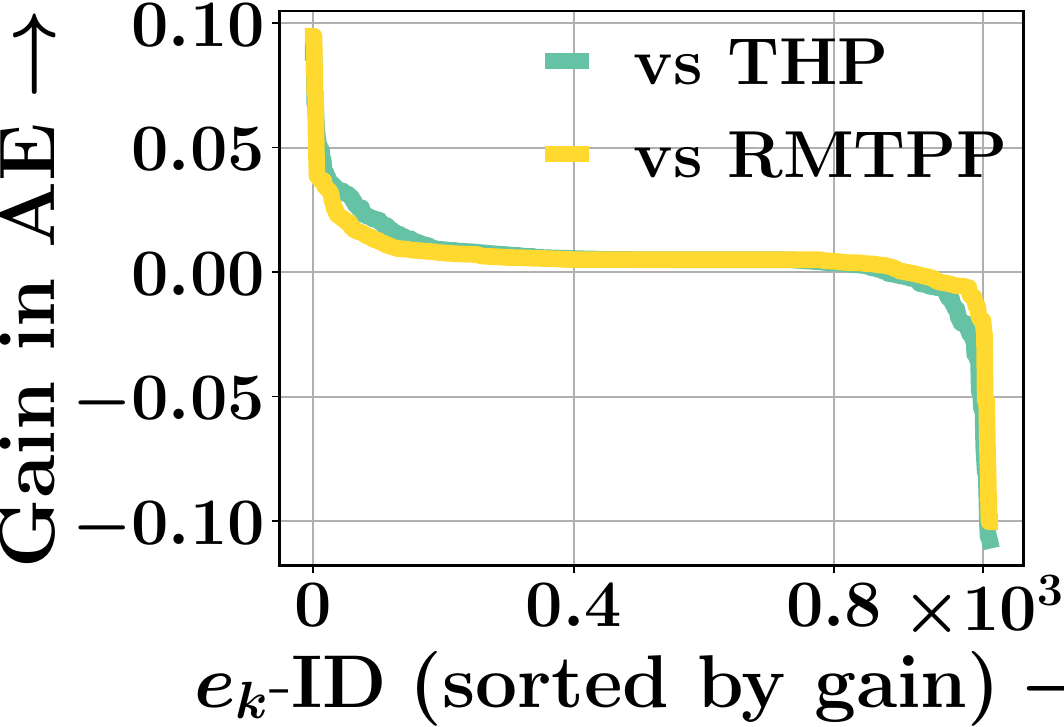}
\caption{\bfast}
\end{subfigure}
\hfill
\begin{subfigure}[b]{0.30\columnwidth}
\centering
\includegraphics[height=2.6cm]{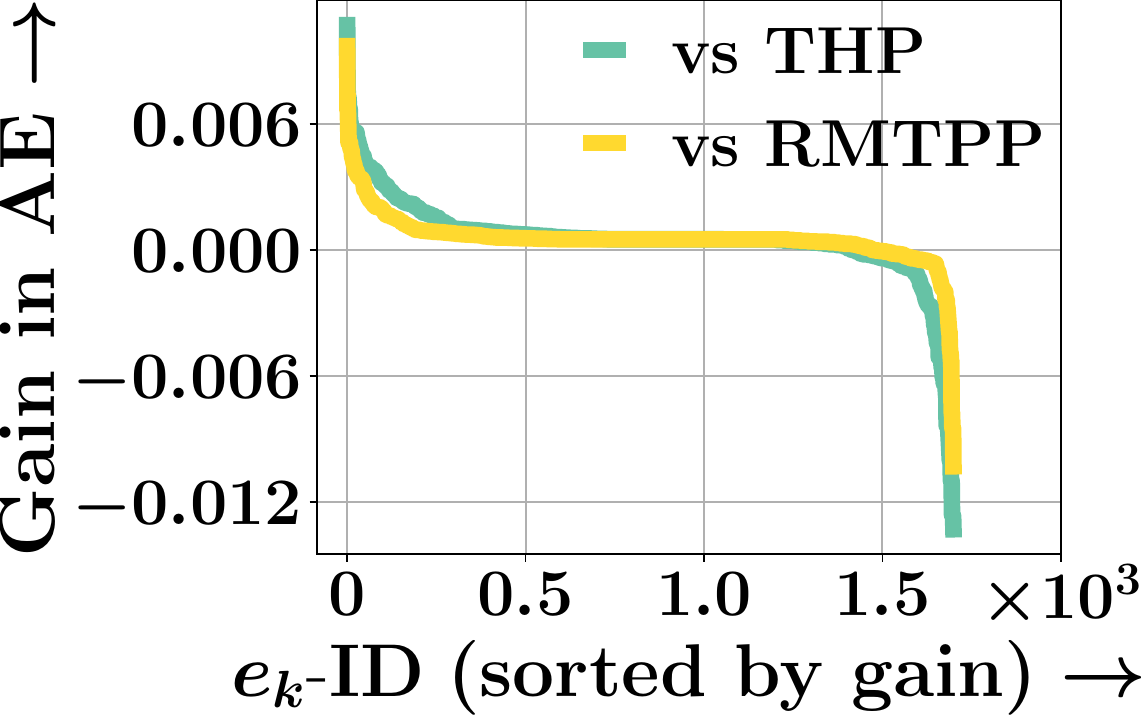}
\caption{\mult}
\end{subfigure}
\hfill
\begin{subfigure}[b]{0.3\columnwidth}
\centering
\includegraphics[height=2.7cm]{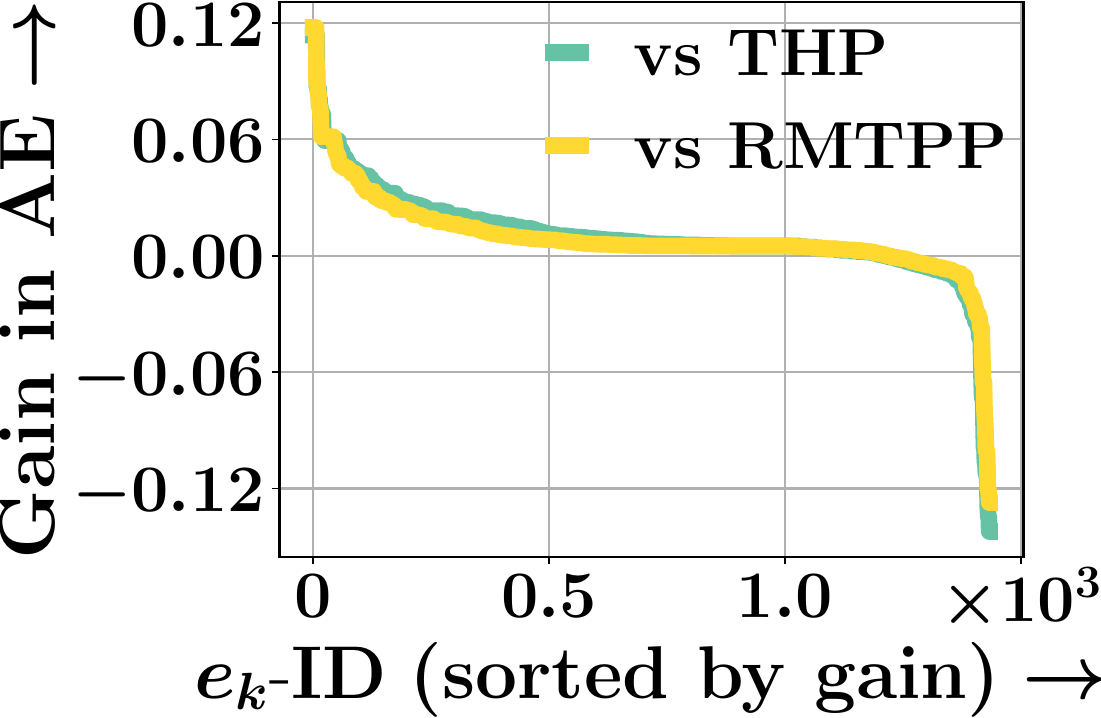}
\caption{\act}
\end{subfigure}
\hfill
\vspace{-0.3cm}
\caption{\label{fig:drilldown} Performance gain in terms of $\text{AE}(\text{baseline})-\text{AE}(\ourm)$ --- the gain (above x-axis) or loss (below the x-axis) with respect to RMTPP and THP. Events are sorted by decreasing gain of \ourm along $x$-axis.\vspace{-0.5cm} }
\end{figure}

\subsection{Drill-Down Analysis}
In this section, we conduct a comparative analysis of time prediction performance at the level of each action in the test set. For every action $e_i$ in the test set, we calculate the difference in time prediction error per action, denoted as $\mathbb{E}[|t_k-\hat{t_k}|]$, between \ourm and two competitive baselines, such as RMTPP and THP, for all datasets. The results, summarized in Figure~\ref{fig:drilldown}, demonstrate that \ourm consistently outperforms the most competitive baseline, RMTPP, for over 75\% of actions across all datasets. Moreover, the performance gain of \ourm over THP is even more pronounced, highlighting its exceptional performance in time prediction accuracy.

\begin{figure}[t]
\centering
\hfill
\begin{subfigure}[b]{0.30\columnwidth}
\centering
\includegraphics[height=3cm]{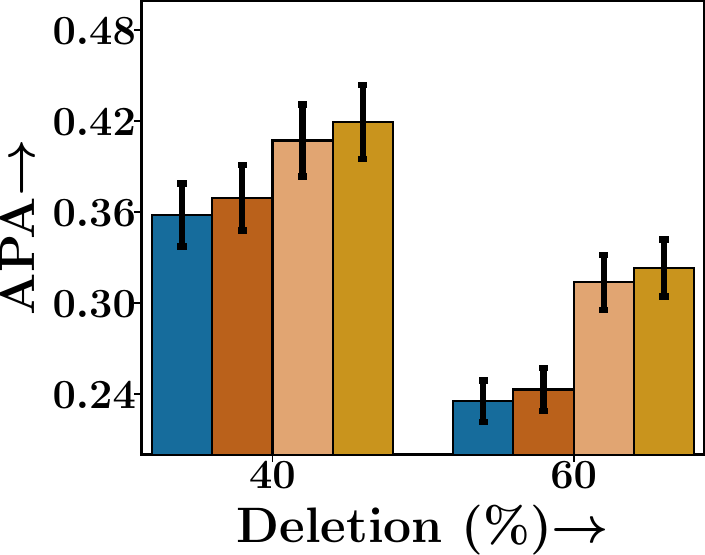}
\caption{\bfast\ (APA)}
\end{subfigure}
\hfill
\begin{subfigure}[b]{0.30\columnwidth}
\centering
\includegraphics[height=3cm]{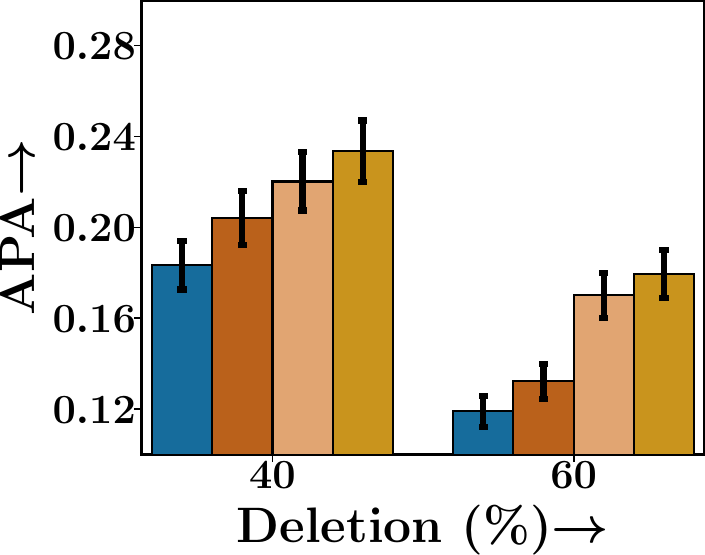}
\caption{\mult\ (APA)}
\end{subfigure}
\hfill
\begin{subfigure}[b]{0.3\columnwidth}
\centering
\includegraphics[height=3cm]{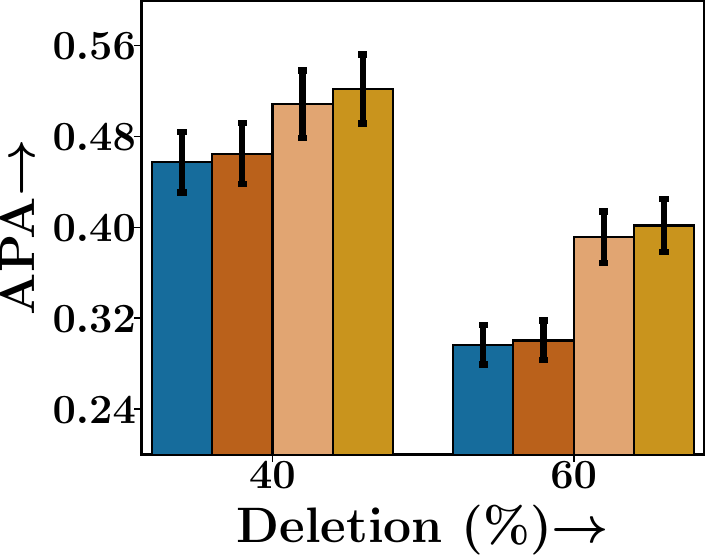}
\caption{\act\ (APA)}
\end{subfigure}
\hfill
\vspace{0.4cm}

\begin{subfigure}[b]{0.30\columnwidth}
\centering
\includegraphics[height=3cm]{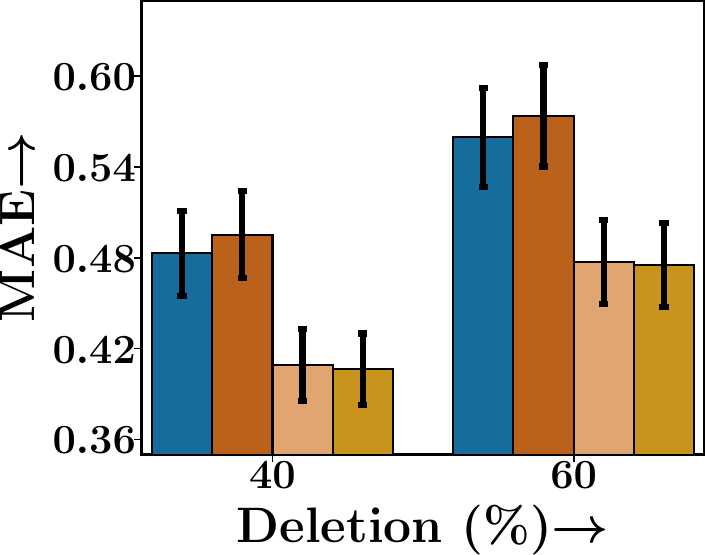}
\caption{\bfast\ (MAE)}
\end{subfigure}
\hfill
\begin{subfigure}[b]{0.30\columnwidth}
\centering
\includegraphics[height=3cm]{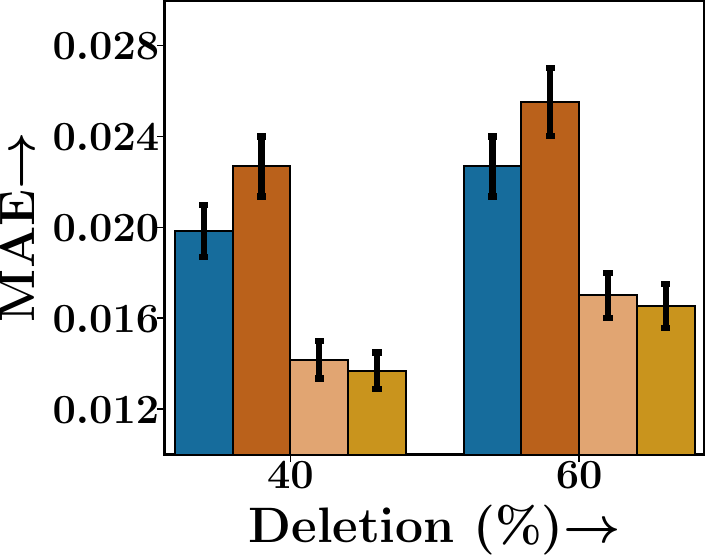}
\caption{\mult\ (MAE)}
\end{subfigure}
\hfill
\begin{subfigure}[b]{0.3\columnwidth}
\centering
\includegraphics[height=3cm]{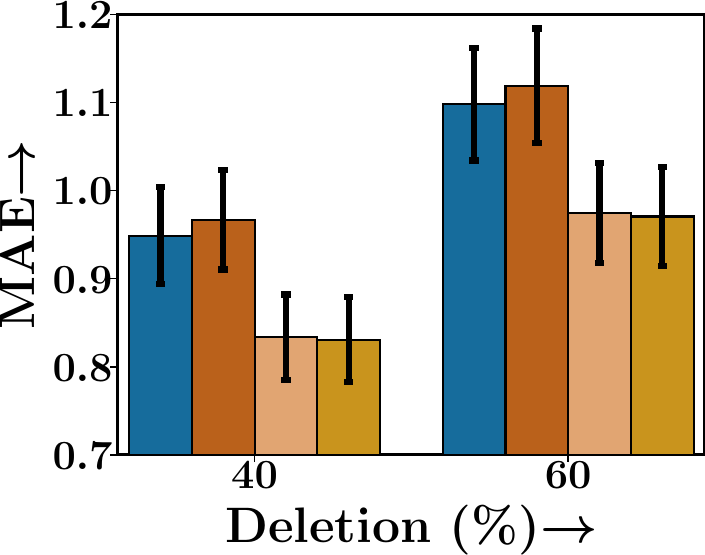}
\caption{\act\ (MAE)}
\end{subfigure}

{\includegraphics[height=0.5cm]{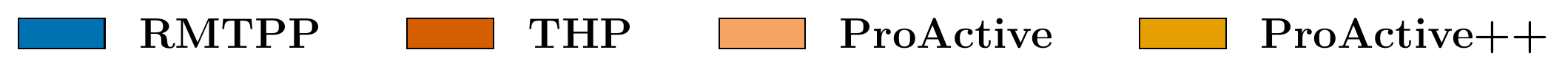}}
\vspace{-0.3cm}
\caption{\label{fig:missing} Action prediction performance in the presence of synthetically deleted events. The results are reported for \bfast, \mult, and \act datasets, in terms of APA and MAE. \vspace{-0.6cm}}
\end{figure}

\subsection{Performance with Missing Data}
To emphasize the applicability of \ourm in the presence of missing data, we perform action prediction on sequences with limited training data. Specifically, we synthetically delete actions \textit{randomly} from a sequence, \ie, we randomly delete 40\% (and 60\%) of actions from the original sequence using a normal distribution. We then train and test our model on the remaining 60\% ( and 40\%) of actions. Figure~\ref{fig:missing} summarizes the results, and we observe that even after a significant quantity of actions are deleted with synthetic data deletion, \ourp shows a significant performance improvement over the best-performing baselines, namely RMTPP and THP. This is because \ourm and \ourp are trained to capture the relationship between the sequence goal and the actions leading to the goal. Consequently, they can effectively exploit the underlying setting with data deletion compared to other models. However, the performance gains saturate with a further increase in missing data, as the added noise in the datasets severely hampers the learning of both models. Interestingly, we note that \ourp outperforms \ourm. This could be attributed to the ability of \ourp to capture the goal without considering the order in which actions occur, allowing it to better correlate each action with the sequence goals. This finding further reinforces the importance of \ourp.

\begin{figure*}[t]
\centering

\begin{subfigure}[b]{\linewidth}
\centering
{\includegraphics[height=3.2cm]{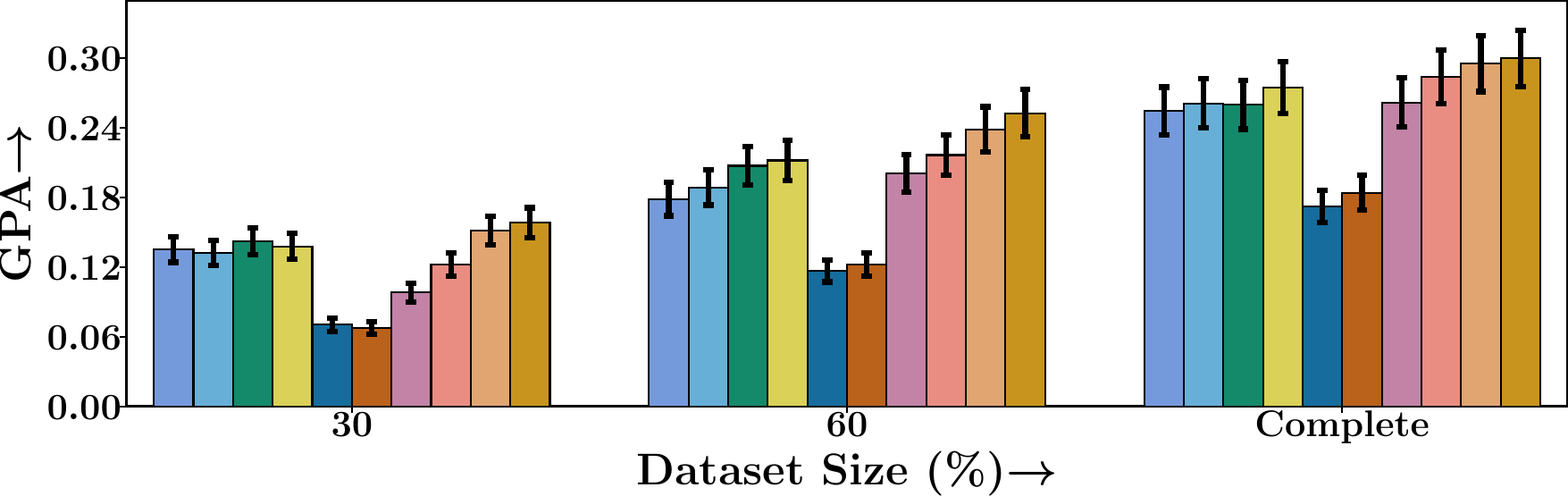}}
\caption{\bfast \vspace{0.2cm}}
\end{subfigure}

\begin{subfigure}[b]{\linewidth}
\centering
{\includegraphics[height=3.2cm]{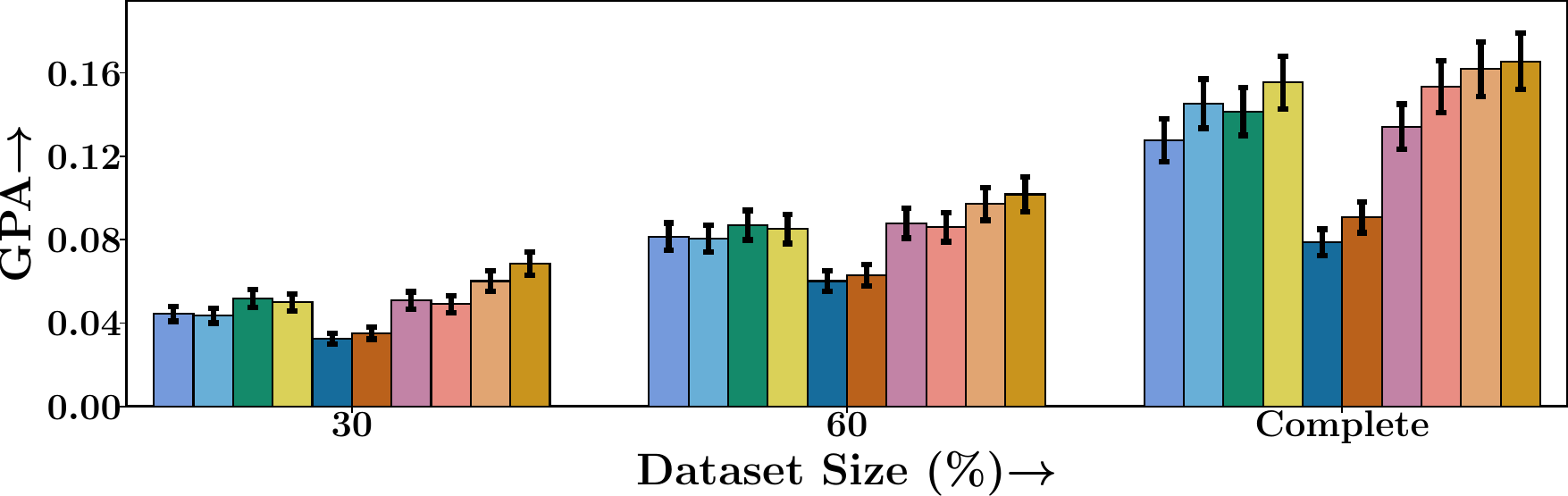}}
\caption{\mult \vspace{0.2cm}}
\end{subfigure}

\begin{subfigure}[b]{\linewidth}
\centering
{\includegraphics[height=3.2cm]{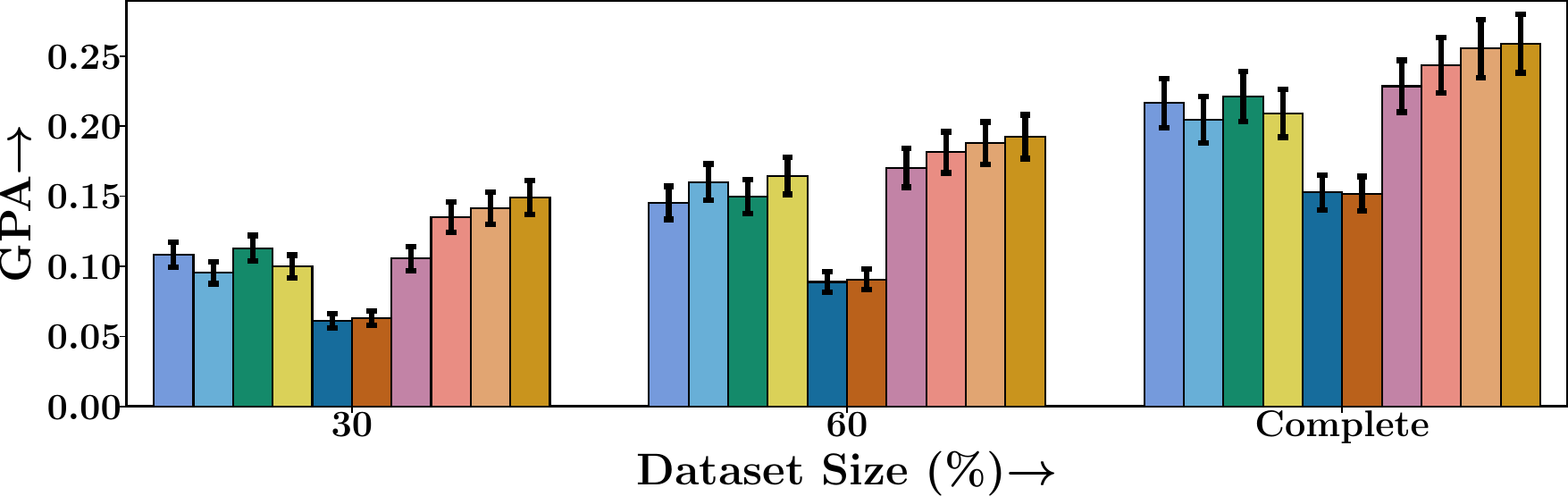}}
\caption{\act \vspace{0.2cm}}
\end{subfigure}

\begin{subfigure}[b]{\linewidth}
\centering
{\includegraphics[height=3.2cm]{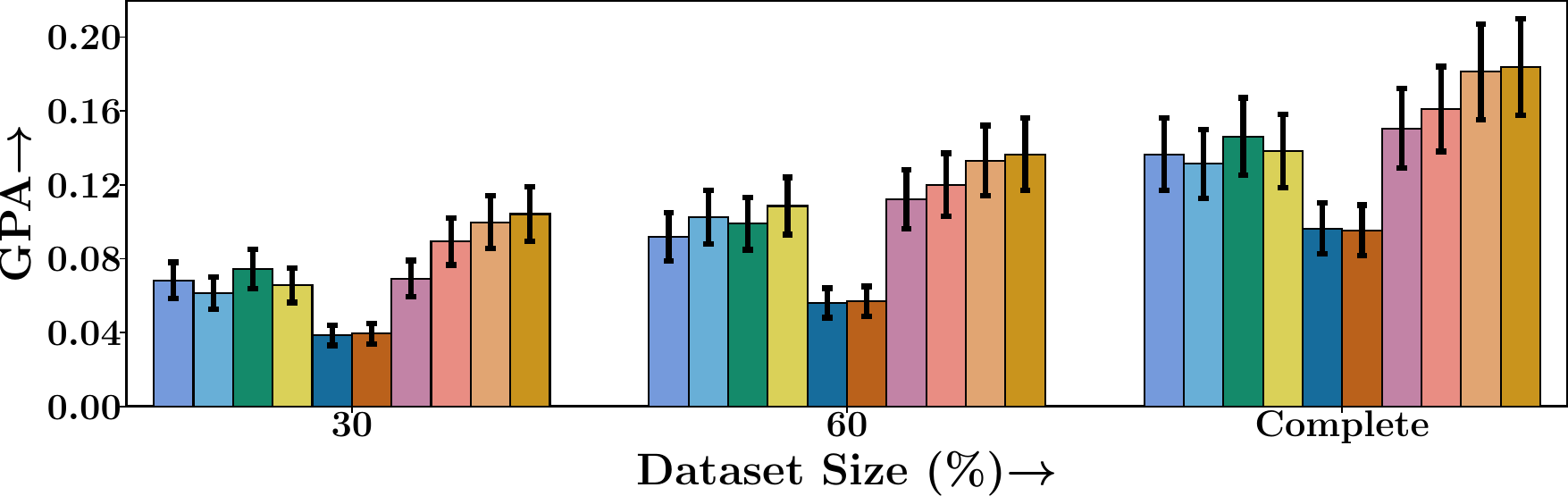}}
\caption{\actnew \vspace{0.2cm}}
\end{subfigure}

{\includegraphics[height=0.8cm]{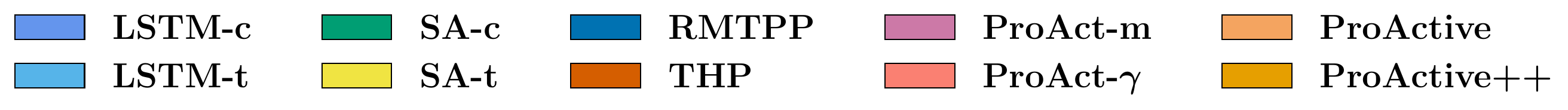}}
\vspace{-0.3cm}
\caption{Sequence goal prediction performance of \ourm, its variants -- \ours-m and \ours-$\gamma$, and other baseline models. The results show that \ourm can effectively detect the CTAS goal even with smaller test sequences as input. \vspace{-0.3cm}}
\label{fig:UU}
\end{figure*}

\subsection{Predicting the Sequence Goal (RQ2)}
Here, we evaluate the goal detection performance of \ourm along with other baselines. To highlight the \textit{early} goal detection ability of our model,  we report the results across different variants of the test set, \ie, with the initial 30\% and 60\% of the actions in the CTAS in terms of goal prediction accuracy (GPA). In addition, we introduce two novel baselines, LSTM-c, and LSTM-t, that detect the CTAS goal using just the types and the times of actions respectively. Similarly, we use self-attention based approaches, SA-c and SA-t. We also compare with the two best-performing MTPP baselines -- RMTPP and THP which we extend for the task of goal detection by a $k$-means clustering algorithm. In detail, we obtain the sequence embedding, say $\bs{s}_k$ using the MTPP models and then cluster them into $|\cm{G}|$ clusters based on their cosine similarities and perform a maximum \textit{polling} across each cluster, \ie, predict the most common goal for each cluster as the goal for all CTAS in the same cluster. In addition, we introduce two new variants of our model to analyze the benefits of early goal detection procedures in \ourm -- (i) \ourm-m, represents our model without the goal-based margin loss given in Eqn.~\eqref{eqn:margin} and (ii) \ourm-$\gamma$, is our model without the discount-factor weight in Eqn.~\eqref{eqn:discount}. We also report the results for the complete model \ourm.

The results for goal detection in Figure~\ref{fig:UU}, show that the complete design of \ourm achieves the best performance among all other models. We also note that the performance of MTPP-based models deteriorates significantly for this new task which shows the unilateral nature of the prediction prowess of MTPP models, unlike \ourm. Interestingly, the variant of \ourm without the margin loss \ourm-m performs poorly as compared to the one without the discount factor, \ourm-$\gamma$. This could be attributed to better convergence guarantees with a margin-based loss over the latter. Finally, we observe that standard LSTM models are easily outperformed by our model, thus reinforcing the need for joint training of types and action times.

\begin{figure}[t]
\centering
\hfill
\begin{subfigure}[b]{0.24\columnwidth}
\includegraphics[height=3cm]{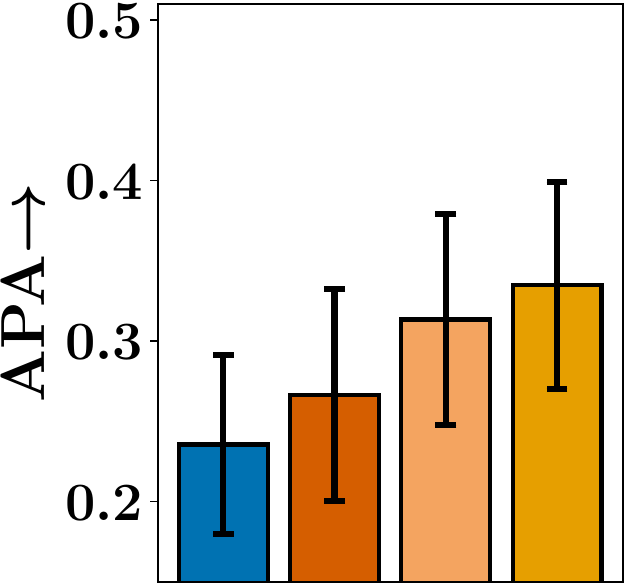}
\caption{\bfast}
\end{subfigure}
\hfill
\begin{subfigure}[b]{0.24\columnwidth}
\includegraphics[height=3cm]{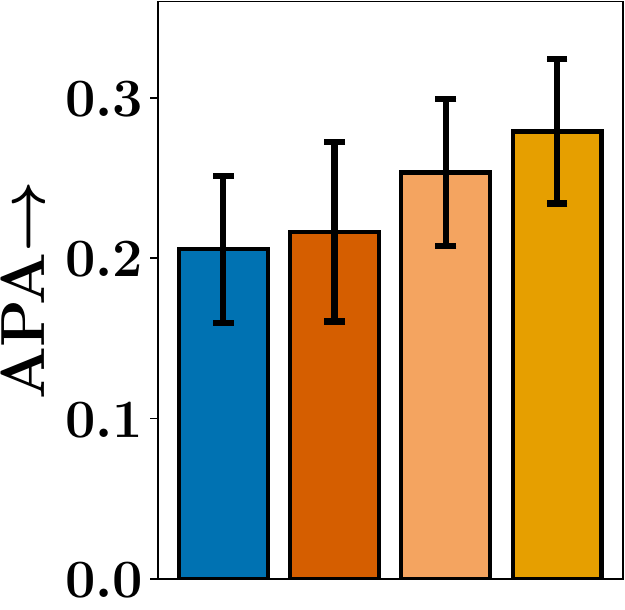}
\caption{\mult}
\end{subfigure}
\hfill
\begin{subfigure}[b]{0.24\columnwidth}
\includegraphics[height=3cm]{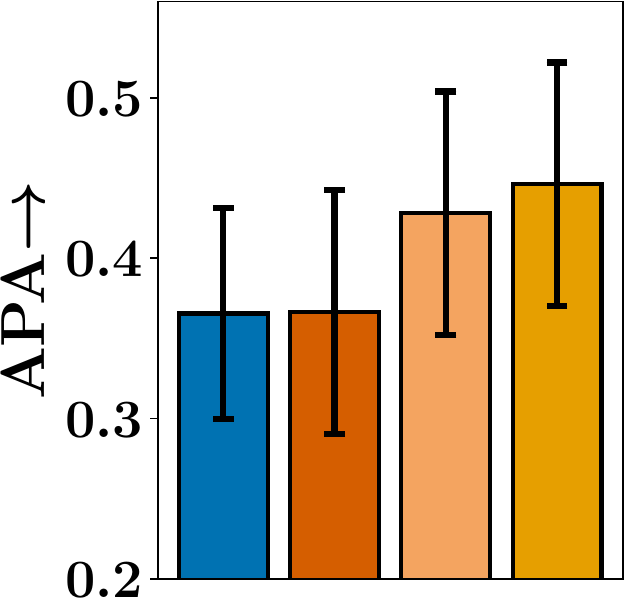}
\caption{\act}
\end{subfigure}
\hfill
\begin{subfigure}[b]{0.24\columnwidth}
\includegraphics[height=3cm]{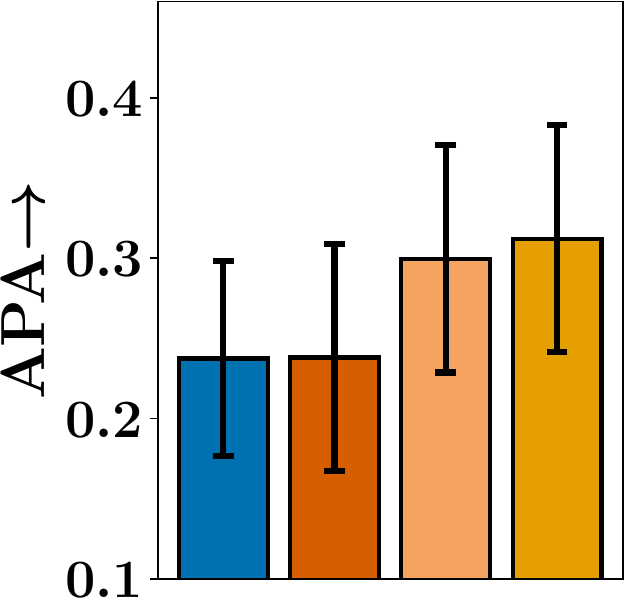}
\caption{\actnew}
\end{subfigure}
\hfill
{\includegraphics[height=0.5cm]{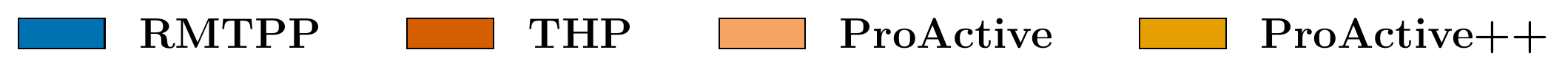}}
\vspace{-0.3cm}
\caption{\label{fig:mpa} Sequence Generation results for \ourm and other baselines in terms of APA for action prediction. \vspace{0.3cm}}
\end{figure}

\begin{figure}[t]
\centering
\hfill
\begin{subfigure}[b]{0.24\columnwidth}
\includegraphics[height=2.8cm]{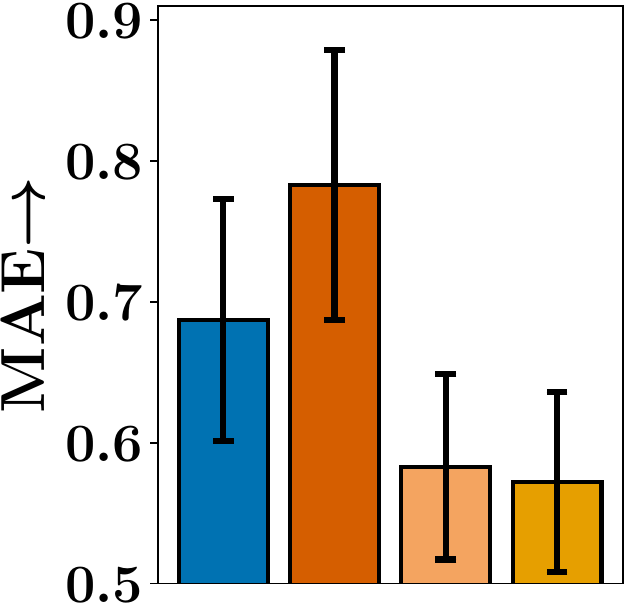}
\caption{\bfast}
\end{subfigure}
\hfill
\begin{subfigure}[b]{0.24\columnwidth}
\includegraphics[height=2.8cm]{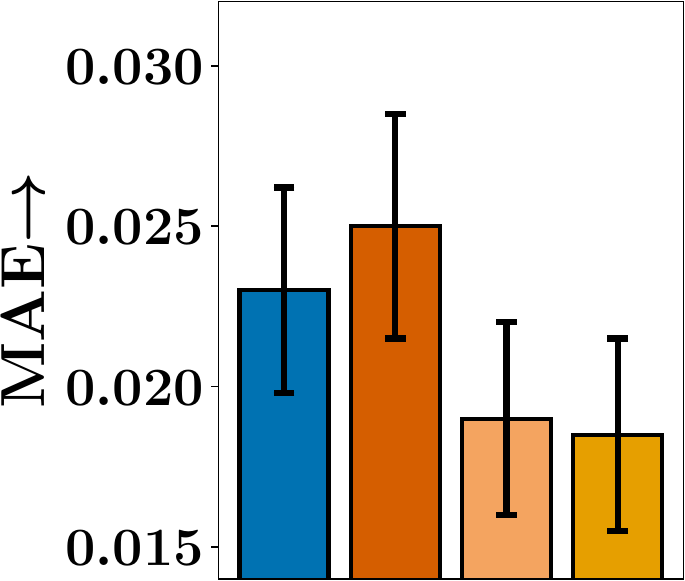}
\caption{\mult}
\end{subfigure}
\hfill
\begin{subfigure}[b]{0.24\columnwidth}
\includegraphics[height=2.8cm]{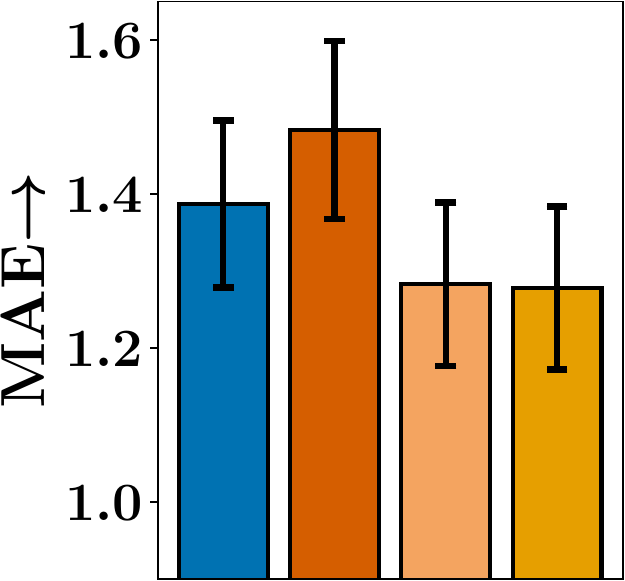}
\caption{\act}
\end{subfigure}
\hfill
\begin{subfigure}[b]{0.24\columnwidth}
\includegraphics[height=2.8cm]{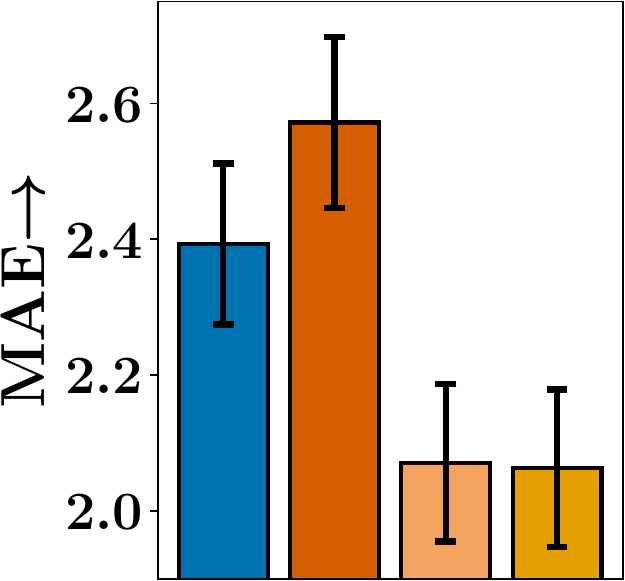}
\caption{\actnew}
\end{subfigure}
\hfill
{\includegraphics[height=0.5cm]{figs/legend_gen.pdf}}
\vspace{-0.3cm}
\caption{\label{fig:mae} Sequence Generation results for \ourm and other baselines in terms of MAE for time prediction.}
\end{figure}

\subsection{Sequence Generation (RQ3)}
Here, we evaluate the sequence generation ability of \ourm. Specifically, we generate all the sequences in the test set by giving the \textit{true} goal of the CTAS and the first action as input to the procedure described in Section~\ref{sec:generation}. However, there may be differences in the lengths of the generated and true sequences, \ie, the length of generated sequences is usually greater than the true CTAS. Therefore, we compare the actions in the true sequence with the initial $|\cm{S}|$ generated actions. Such an evaluation procedure provides us the flexibility of comparing with other MTPP models such as RMTPP~\cite{rmtpp} and THP~\cite{thp}. As these models cannot be used for end-to-end sequence generation, we alter their underlying model for \textit{forecasting} future actions given the first action and then incrementally update and sample from the MTPP parameters. We report the results in terms of APA and MAE for action and time prediction in Figure~\ref{fig:mpa} and Figure~\ref{fig:mae} respectively. The results show that \ourm can better capture the generative dynamics of a CTAS in comparison to other MTPP models. We also note that the prediction performance deteriorates significantly in comparison to the results given in Table~\ref{tab:apa} and Table~\ref{tab:mae}. This could be attributed to the error that gets compounded in the further predictions made by the model. Interestingly, the performance advantage that \ourm has over THP and RMTPP is further widened during sequence generation. 

\begin{figure}[t]
\centering
\hfill
\begin{subfigure}[b]{0.24\columnwidth}
\includegraphics[height=2.8cm]{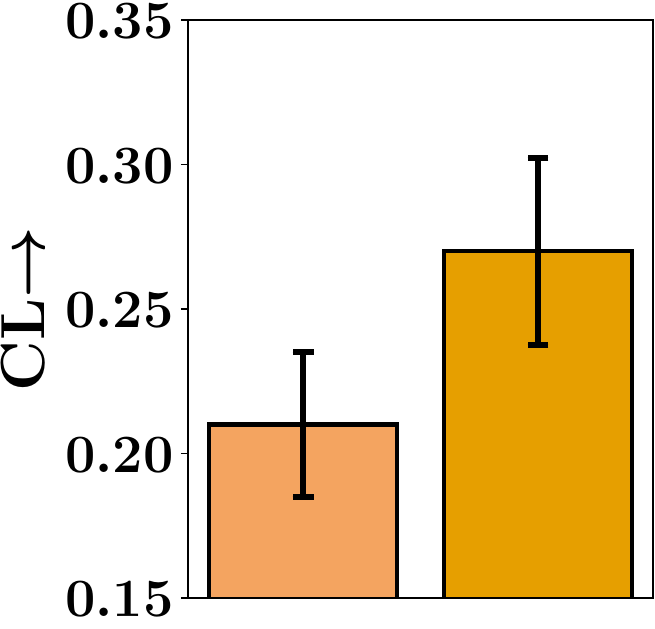}
\caption{\bfast}
\end{subfigure}
\hfill
\begin{subfigure}[b]{0.24\columnwidth}
\includegraphics[height=2.8cm]{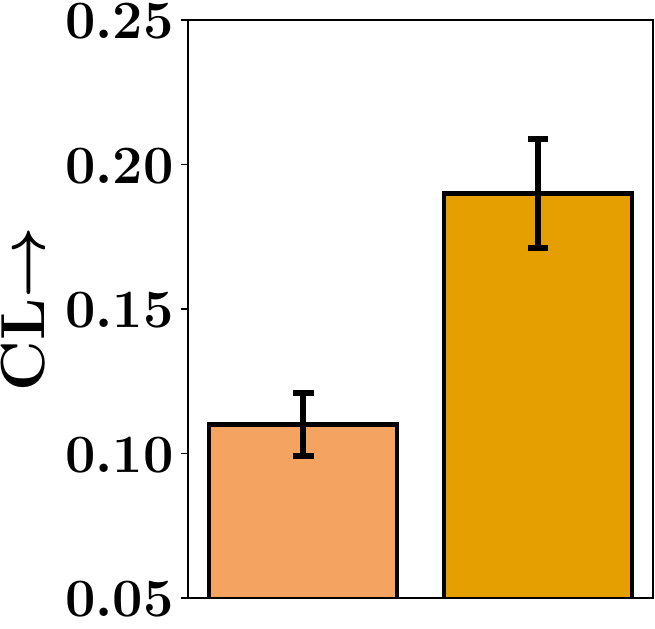}
\caption{\mult}
\end{subfigure}
\hfill
\begin{subfigure}[b]{0.24\columnwidth}
\includegraphics[height=2.8cm]{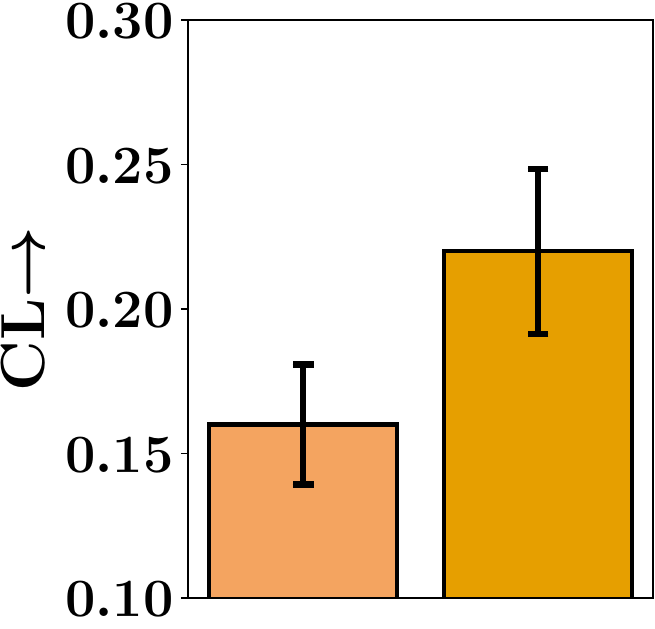}
\caption{\act}
\end{subfigure}
\hfill
\begin{subfigure}[b]{0.24\columnwidth}
\includegraphics[height=2.8cm]{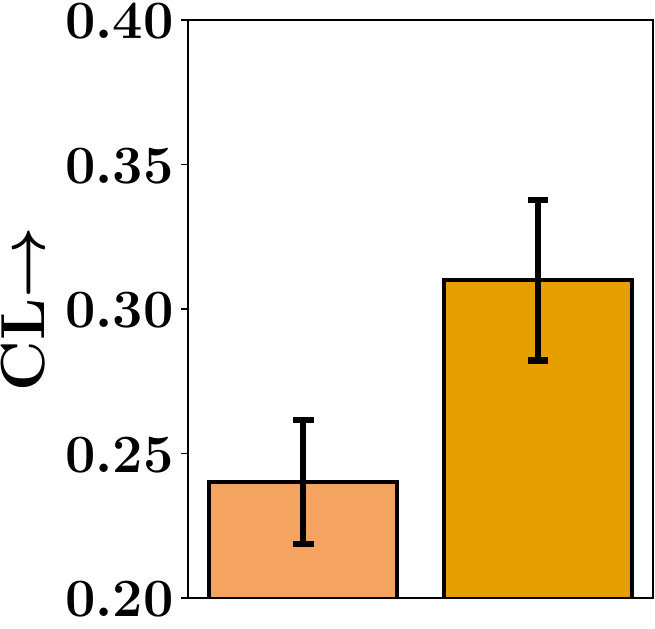}
\caption{\actnew}
\end{subfigure}
\hfill
{\includegraphics[height=0.5cm]{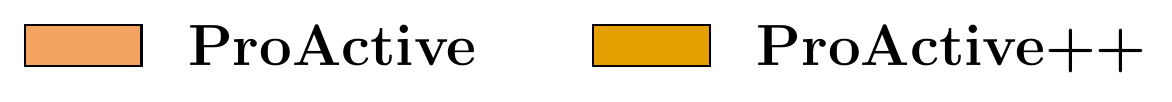}}
\vspace{-0.3cm}
\caption{\label{fig:cl} Comparing the Length of Sequences generated by \ourm, \ourp and the true CTAS. \vspace{-0.3cm}}
\end{figure}

\xhdr{Length Comparison}
Here, we conduct a length comparison analysis to evaluate the effectiveness of our generated sequences compared to the true sequences. The purpose was to assess how well our model, referred to as \ourm, captured the generative mechanism of the sequences. The length comparison was quantified using a metric called the Correct-Length ratio (CL), which was calculated as follows:
\begin{equation}
\mathrm{CL} = \frac{1}{N}\sum_{\forall \cm{S}} \#(|\cm{S}|=|\widehat{\cm{S}}|),
\end{equation}
Here, $N$ represents the total number of sequences in the dataset. For each sequence $\cm{S}$ in the dataset, we checked if the length of the generated sequence $\widehat{\cm{S}}$ was equal to the length of the true sequence $|\cm{S}|$. The symbol $\#$ denotes the count of instances where the lengths were equal. The average of these counts, divided by $N$, yielded the CL value. For our experimental datasets, the obtained CL values are reported in Figure~\ref{fig:cl}. These values indicate the percentage of instances where \ourm successfully generated sequences with the correct length. At first glance, these results might appear substandard. However, considering the inherent difficulty associated with the task of sequence generation solely based on the CTAS goal, we consider these values to be satisfactory. We believe that the generation procedure employed by \ourm opens up new possibilities and avenues for generating action sequences. In conclusion, while the CL values obtained may not be exceptionally high, we consider our results to be encouraging in the domain of CTAS generation.

\subsection{Parameter Sensitivity (RQ4)}
\begin{figure}[t]
\centering
\hfill
\begin{subfigure}[b]{0.3\columnwidth}
\includegraphics[height=3cm]{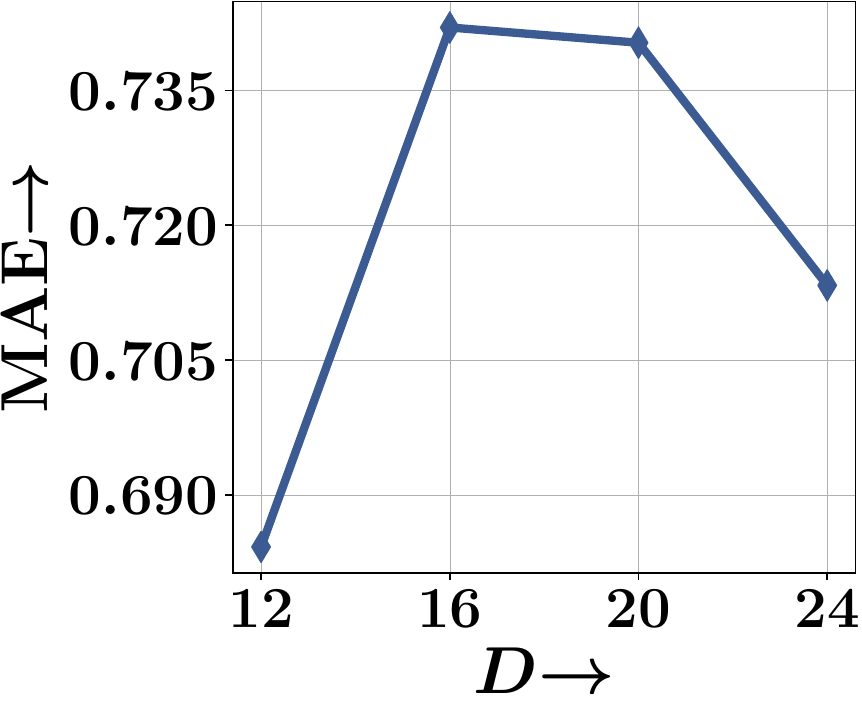}
\caption{Across $D$}
\end{subfigure}
\hfill
\begin{subfigure}[b]{0.3\columnwidth}
\includegraphics[height=3cm]{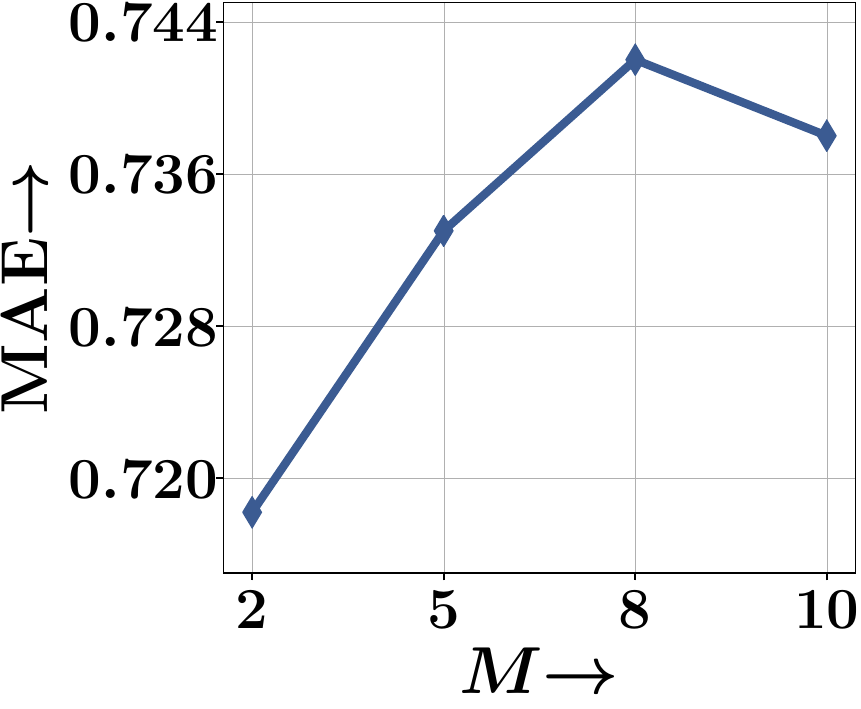}
\caption{Across $\cm{M}$}
\end{subfigure}
\hfill
\begin{subfigure}[b]{0.3\columnwidth}
\includegraphics[height=3cm]{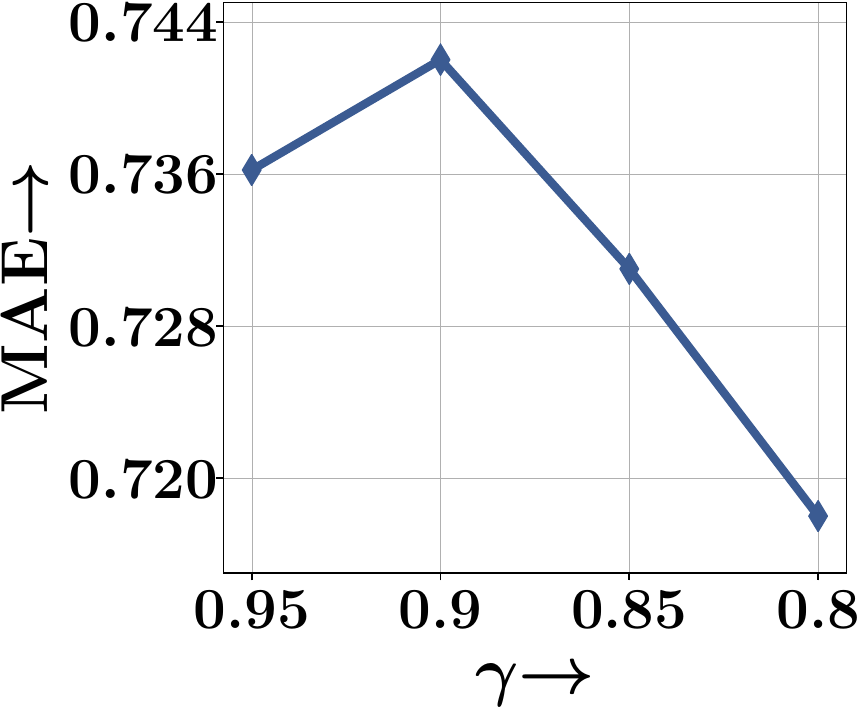}
\caption{Across $\gamma$}
\end{subfigure}
\hfill
\vspace{-0.3cm}
\caption{\label{fig:pars} \ourm sensitivity for \act\ dataset with different hyper-parameter values. \vspace{-0.3cm}}
\end{figure}

Finally, we perform the sensitivity analysis of \ourm over key parameters: (i) $D$, the dimension of embeddings; (ii) $\cm{M}$, no. clusters for lognormal flow; and (iii) $\gamma$, the discount factor described in Eqn.~\eqref{eqn:discount}. For brevity purposes, we only report results on the \act dataset, but other datasets displayed similar behavior. From Figure~\ref{fig:pars} we show the performance of \ourm across different hyper-parameter values. We note that as we increase the embedding dimension the performance first increases since it leads to better modeling. However, beyond a point, the complexity of the model increases requiring more training to achieve good results, and hence we see its performance deteriorating. We see a similar trend for $\cm{M}$, as increasing the number of clusters leads to better results before saturating at a certain point. We found $\cm{M}$=5 to be the optimal point across datasets in our experiments. Finally, across $\gamma$, we notice that smaller values for gamma penalize the loss function for detecting the goal late, however, it deteriorates the action prediction performance of \ourm. Therefore, we found $\gamma$=0.9 as the best trade-off between goal and action prediction.

\subsection{Scalability} 
We found that the training run-times for \ourm and \ourp on all datasets were within 1 hour. This observation is significant because it indicates that the training process of our model is relatively fast and practical for deployment in real-world scenarios. Furthermore, the relatively fast training times of \ourm align with our design choice of utilizing a neural MTPP for modeling a continuous-time action sequence. The decision to employ neural point process flows was motivated by its faster learning capabilities and closed-form sampling~\cite{intfree,ppflows,vinayak_thesis}. This enables the model to generate sequences that exhibit coherent and meaningful temporal dynamics. Additionally, the closed-form sampling technique employed by \ourm allows for the efficient and straightforward generation of sequences, further contributing to its practicality.

\section{Related Work} \label{sec:related}
In this section, we introduce key related work for this paper. It mainly falls into -- activity prediction and temporal point processes.

\subsection{Activity Prediction using Visual or Sequence Frameworks}
Activity modeling in videos is a widely used application with recent approaches focusing on frame-based prediction. \citet{conc1} predicts the future actions via hierarchical representations of short clips, \citet{conc2} jointly predicts future activity and the starting time by capturing different sequence features. A similar procedure is adopted by \cite{conc3} that predicts the action categories of a sequence of future activities as well as their starting and ending times. \citet{margin} propose a method for early classification of a sequence of frames extracted from a video by maximizing the margin-based loss between the correct and the incorrect categories, however, it is limited to visual data and cannot incorporate the action-times. This limits its ability for use in CTAS, and sequence generation. A recent approach~\cite{avae} proposed to model the dynamics of action sequences using a variational auto-encoder (VAE) built on top of a temporal point process. We consider their work as most relevant to \ourm as it also addressed the problem of CTAS modeling. However, as shown in our experiments \ourm was able to easily outperform it across all metrics. This could be attributed to the limited modeling capacity of VAE over normalizing flows. Moreover, their sampling procedure could not be extended to sequence generation. Therefore, in contrast to the past literature, \ourm is the first application of MTPP models for CTAS modeling and end-to-end sequence generation.

\subsection{Marked Temporal Point Processes for Event Sequences}
In recent years neural Marked Temporal Point Processes (MTPP) have shown significant promise in modeling a variety of continuous-time sequences in healthcare~\cite{rizoiu2}, finance~\cite{sahp,bacry}, education~\cite{sahebi}, and social networks~\cite{leskovec,colab,imtpp,proactive,reformd}. However, due to the limitations of traditional MTPP models, in recent years, neural enhancements to MTPP models have significantly enhanced the predictive power of these models. Specifically, they combine the continuous-time approach from the point process with deep learning approaches and thus, can better capture complex relationships between events. The most popular approaches~\cite{rmtpp,nhp,intfree,sahp,thp} use different methods to model the time- and mark distribution via neural networks. Specifically,~\citet{rmtpp} embeds the event history to a vector representation via a recurrent encoder that updates its state after parsing each event in a sequence;~\citet{nhp} modified the LSTM architecture to employ a continuous-time state evolution;~\citet{intfree} replaced the intensity function with a mixture of \textit{lognormal} flows for closed-form sampling;~\citet{sahp} utilized the transformer architecture~\citet{transformer} to capture the long-term dependencies between events in the history embedding and~\citet{thp} used the transformer architecture for sequence embedding but extended it to graph settings as well. However, these models are not designed to capture the generative distribution of future events in human-generated sequences.

\section{Conclusion} \label{sec:conc}
Standard deep-learning models are not designed for modeling sequences of actions localized in continuous time. However, neural MTPP models overcome this drawback but have limited ability to model the events performed by a human. Therefore, in this paper, we developed a novel point process flow-based architecture called \ourm for modeling the dynamics of a CTAS. \ourm solves the problems associated with action prediction, CTAS goal detection, and for the first time, we extend MTPP for end-to-end CTAS generation. Our experiments on three large-scale diverse datasets reveal that \ourm can significantly improve over the state-of-the-art baselines across all metrics. Moreover, the results also reinforce the novel ability of \ourm to generate a CTAS. We hope that such an application will open many horizons for using MTPP in a wide range of tasks. As a future work, we plan to incorporate a generative adversarial network~\cite{gan,wgantpp} with action sampling and train the generator and the MTPP model simultaneously.

\begin{acks}
This work was partially supported by a DS Chair of AI fellowship to Srikanta Bedathur. 
\end{acks}

\bibliographystyle{ACM-Reference-Format}
\bibliography{refs}
\end{document}